\documentclass[runningheads]{llncs}

\usepackage{eccv}

\usepackage{eccvabbrv}

\usepackage{graphicx}
\usepackage{booktabs}

\usepackage{amsmath,amssymb}
\usepackage[accsupp]{axessibility}  

\usepackage{hyperref}

\usepackage{orcidlink}

\begin{document}

\title{Explicit Logic Channel for Validation and Enhancement of MLLMs on Zero-Shot Tasks} 

\titlerunning{ELC for MLLM Validation and Enhancement}


\author{Mei Chee Leong \and
Ying Gu \and Hui Li Tan \and Liyuan Li \and Nancy F. Chen}

\authorrunning{M.C. Leong, Y. Gu, H.L. Tan, L. Li, N.F. Chen}

\institute{Institute for Infocomm Research (I$^2$R),\\
\textbf{Present address:}
Institute of Advanced Intelligence and Computing (IAIC),\\
Agency for Science, Technology and Research (A*STAR), Singapore\\
}
\maketitle

\begin{abstract}
Frontier Multimodal Large Language Models (MLLMs) exhibit remarkable capabilities in Visual-Language Comprehension (VLC) tasks. However, they are often deployed as zero-shot solution to new tasks in a black-box manner. Validating and understanding the behavior of these models become important for application to new task. We propose an Explicit Logic Channel, in parallel with the black-box model channel, to perform explicit logical reasoning for model validation, selection and enhancement. The frontier MLLM, encapsulating latent vision-language knowledge, can be considered as an Implicit Logic Channel. The proposed Explicit Logic Channel, mimicking human logical reasoning, incorporates a LLM, a VFM, and logical reasoning with probabilistic inference for factual, counterfactual, and relational reasoning over the explicit visual evidence. A Consistency Rate (CR) is proposed for cross-channel validation and model selection, even without ground-truth annotations. Additionally, cross-channel integration further improves performance in zero-shot tasks over MLLMs, grounded with explicit visual evidence to enhance trustworthiness. Comprehensive experiments conducted for two representative VLC tasks, \ie, MC-VQA and HC-REC, on three challenging benchmarks, with 11 recent open-source MLLMs from 4 frontier families. Our systematic evaluations demonstrate the effectiveness of proposed ELC and CR for model validation, selection and improvement on MLLMs with enhanced explainability and trustworthiness.
\end{abstract}

\section{Introduction}
\label{sec:intro}

\begin{figure}[tb]
  \centering
  \includegraphics[width=0.9\columnwidth]{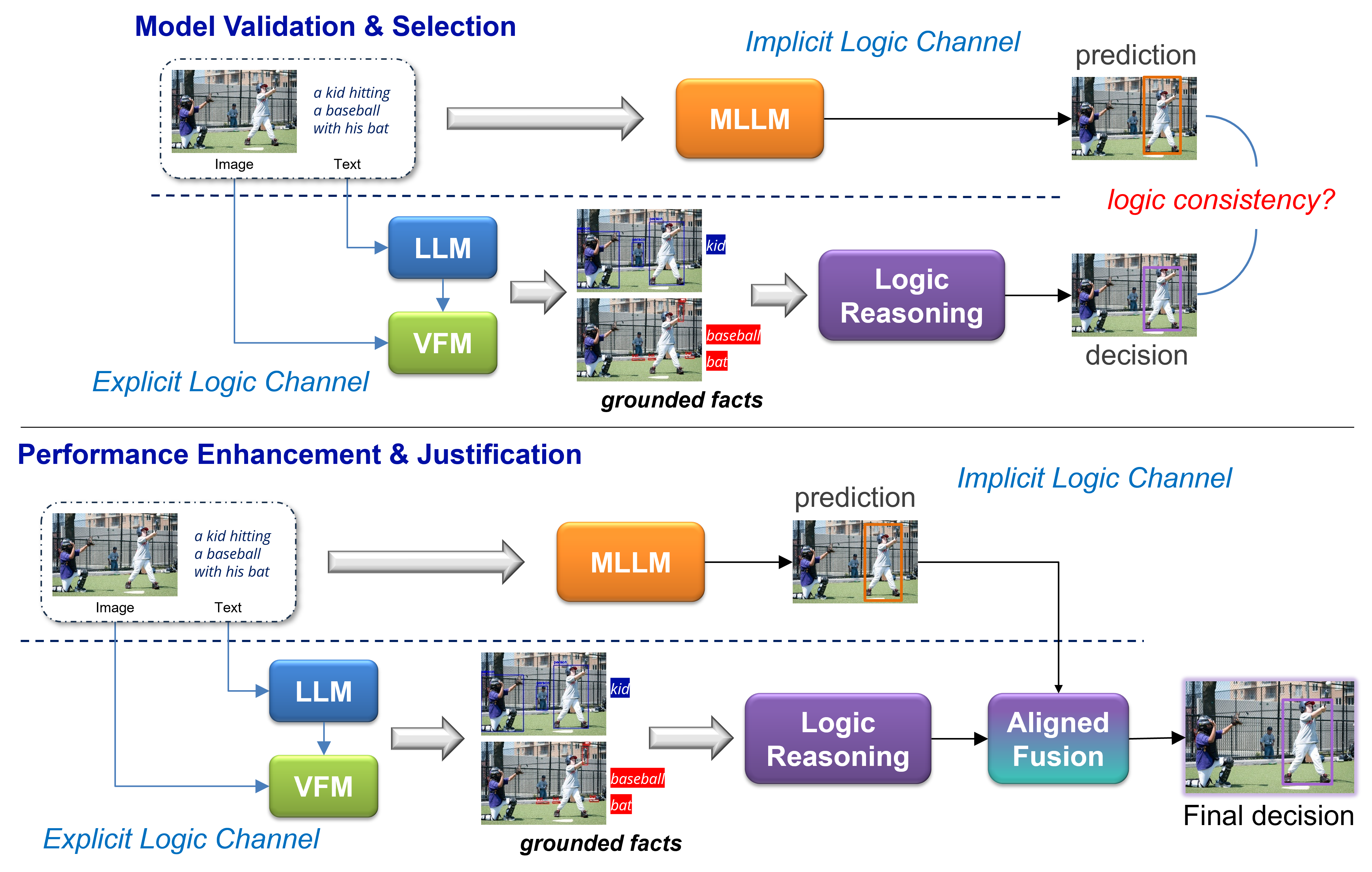}
  \caption{Illustration of Explicit Logic Channel (ELC) for MLLM model validation and selection (upper) and performance enhancement (lower) for novel VLC application in zero-shot setting without the need of ground-truth annotation. In ELC, we combine LLM, VFM and Logical reasoning to derive a prediction on explicit and concrete visual evidence and logical validation for the VLC task.}
  \label{fig:dual-channel-framework}
\end{figure}

Recently, MLLMs have advanced significantly, reflected by the frequent release of frontier models from leading AI organizations and the expansion of vision-language (VL) benchmarks in both scale and diversity~\cite{zhao2025surveylargelanguagemodels, fu2025mmecomprehensiveevaluationbenchmark, yue2024mmmu, li2024surveybenchmarksmultimodallarge, Li_2025_CVPR}, as well as numerous  applications~\cite{liu2024mmbenchmultimodalmodelallaround,wang2024comprehensivereviewmultimodallarge}.
Despite these advances, recent research increasingly highlight the limitations of MLLMs in reliability, factuality, explainability, and logical reasoning~\cite{dang2024explainableinterpretablemultimodallarge, bai2025hallucinationmultimodallargelanguage, Huang_2025, wang2024factualitylargelanguagemodels, Rahman_2026, yuan2025mmereasoningcomprehensivebenchmarklogical, kil2025mllmcompbenchcomparativereasoningbenchmark}. Due to data privacy concerns, large model sizes, and the closed-source nature of models, frontier MLLMs are often used as black-boxes for zero-shot inference on new tasks. 
Therefore, it is crucial to be able to identify reliable models and improve prediction accuracy without fine-tuning for new tasks~\cite{Khan_2024_CVPR, zhao2021calibrateuseimprovingfewshot, wang2023selfconsistencyimproveschainthought,zhao2021calibrateuseimprovingfewshot, zhao2023automaticmodelselectionlarge,zhi2025tfar}.

Vision-Language tasks, such as VQA and VG (Visual Grounding), in general, require a model to make a decision to a text query on visual information, needing a high-level understanding and concrete visual evidence to support the response. Recently, significant progresses have been achieved for enhancing the visual-language consistency and reliability, such as Grounded VQA~\cite{7780909,8953451,Chen_2022_CVPR,chen2024lcv2,chen2023vqatheraphy} and VideoQA~\cite{Xiao_2024_CVPR,di2024grounded_ego,li2022invariant}. The advancements can be classified into three categories: Dataset, Training, and Metric. First, new datasets with additional annotations of related visual information are proposed, including attention region~\cite{Chen_2022_CVPR}, object~\cite{chen2023vqatheraphy}, and scene-graph~\cite{DBLP:journals/corr/abs-2509-11862}. Second, extensions of architecture and grounding objective function are proposed to force the model also providing grounded visual cue with the text answer on the new datasets with additional annotations~\cite{9578483,10.1007/978-3-031-19833-5_38}. Third, corresponding to the formats of the extended datasets, new metrics are proposed which evaluate the performance on not only the text answer but also the visual information on the additional annotations~\cite{8953451,reich2022visuallygroundedvqalatticebased,reich-etal-2023-measuring}. These efforts focus on enhancing self-explanation of models on their responses. They might not be applicable for model validation in zero-shot applications without gt annotations. 


As MLLMs are pre-trained on vast VL corpora, they are assumed to have implicitly learned human-like reasoning, forming an Implicit Logic Channel (ILC) that operates as a block-box. 
However, human makes decisions on explicit facts, relationships and logical rules~\cite{10.3389/frai.2022.806403}. 
To validate the correctness of ILC, we introduce a novel Explicit Logic Channel (ELC), that runs in parallel with ILC. In ELC, a LLM is prompted to extract concept-level, task-related facts and relations from the input text. A VFM then grounds these facts explicitly in the image. Novel Logic Reasoning approaches are proposed to perform probabilistic inference on the grounded facts and relations and make the decision with explicit visual evidence. We further propose a Consistency Rate (CR) between the ILC and ELC, enabling model validation, justification, and selection, even without ground-truth (gt) annotations. The explicit visual evidence enhances user trust in predictions, while inconsistent samples are flagged for efficient manual inspection. In addition, based on auto-selected consistent sample set, we propose an aligned fusion to combine both channels for task performance enhancement and model verification without gt annotation and model fine-tuning.

To evaluate the effectiveness of ELC and CR for zero-shot VLC tasks, we conduct experiments on two representative tasks, \ie, MC-VQA and HC-REC, using three recent challenging benchmarks, \ie, NegBench, HC-RefCOCOg and HC-RefLoCo. We examine 11 frontier open-source MLLMs across four leading families, \ie, Gemma, LLaVA, InternVL and QwenVL. Our experimental results show that, (a) the metric CR is strongly correlated with accuracy and provides reliable evaluation without gt annotation; (b) ELC with CR enables effective model validation and selection for VLC tasks without gt annotations; (c) the aligned fusion of ILC and ELC further improves VLC performance while offering explicit logical justification for enhanced trustworthiness.

Our main contributions include: (1) A general and adaptable Explicit Logic Channel with foundation models and Logic Reasoning, enabling validation, selection and enhancement of MLLMs on novel VLC tasks without gt annotation; (2) A logic consistency metric CR for model performance evaluation without requiring ground-truth; (3) A comprehensive study across 11 frontier MLLMs from four leading families, demonstrating the effectiveness of explicit logic consistency analysis for zero-shot VL applications.

\section{Related work}
\label{sec:related_work}

\noindent {\bf LVLMs.} Large Vision-Language Models (LVLMs) have grown rapidly in recent years ~\cite{Li_2025_CVPRW,10.5555/3666122.3666201}. 
Representative VLMs, particularly CLIP cluster models~\cite{pmlr-v139-radford21a,Wang2023EquivariantSF}, and those developed on BLIP~\cite{pmlr-v162-li22n} and ALIGN~\cite{jia2021scalingvisualvisionlanguagerepresentation}, are pre-trained on massive VL corpora. Another major category is MLLMs, which adopt a pre-trained LLM as the backbone~\cite{Yin_2024}. Visual features are extracted by a vision encoder, projected into the LLM's token space, and inserted into the token sequence for auto-regressive prediction. Recent MLLMs include LLaVA~\cite{liu2023visualinstructiontuning}, GPT-4V~\cite{yang2023dawnlmmspreliminaryexplorations}, Gemini~\cite{geminiteam2025geminifamilyhighlycapable}, InternVL~\cite{Chen_2024_CVPR_InternVL}, and Qwen-VL~\cite{bai2023qwenvlversatilevisionlanguagemodel}. In this work, we perform evaluations on 11 frontier MLLMs from four leading families. The open-source models with less 10B parameters make them practical for real-world applications. 

\noindent {\bf VLC benchmarks.} 
Benchmarks for evaluating LVLM performance on VLC tasks have expanded rapidly~\cite{Li_2025_CVPRW,fu2024mmesurveycomprehensivesurveyevaluation}. VLC capabilities are commonly assessed using Visual Question Answering (VQA) and Referring Expression Comprehension (REC) or Visual Grounding. Early VQA datasets typically involve multiple-choice questions or short textual answers, while later benchmarks broaden question types, domain knowledge, and tasks such as mathematical reasoning and chart understanding. Recent efforts also examine dataset bias~\cite{NEURIPS2024_1e69ff56} and negation understanding~\cite{alhamoud2025vision,10.1007/978-3-031-72630-9_3}. 
Many standard VQA datasets have become less challenging for frontier models~\cite{Li_2025_CVPRW} due to extensive pre-training. Thus, we adopt a recent benchmark with negations to evaluate the effectiveness of zero-shot setting. The REC task aims to localize an image region based on a referring expression~\cite{xiao2024visualgroundingsurvey}. Standard datasets include RefCOCO~\cite{yu2016refcoco}, RefCOCO+~\cite{yu2016refcoco}, and RefCOCOg~\cite{mao2016refcocog}, constructed from MS COCO. However, RefCOCO/+ might not be adequate for evaluating LVLM due to their concise referring phrases and limited vocabulary. On the other hand, zero-shot REC remains challenging~\cite{han2024zerorec}. Therefore, we conduct experiments on HC-RefCOCOg, which features richer descriptions (8.9 average words vs 3.3 and 3.4 in RefCOCO/+), and HC-RefLoCo~\cite{NEURIPS2024_80f0cd03}, a recent challenge with long context expression of 93 words in average.

\noindent {\bf Logic reasoning on LLMs and VLM.} Enhancing the logical reasoning capabilities of LLMs have attracted research attention~\cite{cheng2025empoweringllmslogicalreasoning,pournemat2025reasoninguncertaintyexploringprobabilistic}. Prior work focuses on improving logical accuracy and consistency in generated responses. The approaches include introducing an external logic solver~\cite{olausson-etal-2023-linc,calanzone2025logically} or building additional dataset with logically consistent annotations to fine-tune a model~\cite{feng-etal-2024-language}. In BIRD~\cite{feng2025bird}, probabilistic inference is used to improve the trustworthiness of LLM. Existing NeuroSymbolic frameworks are sequential structures from Neuro-Networks to Logic Engines, and the programming methods typically require additional learning to perform logic reasoning~\cite{ZiyangLi2023,Huang2023LASERAN,Yang2024_10234506,li2023scalloplanguageneurosymbolicprogramming}.
Different from these approaches, we propose a Dual-Channel framework and the ELC performs logic reasoning on grounded facts without additional training. 


\section{Methodology}
\label{sec:method}
A VLC task can be defined as follows: given an image $I$ and a text expression $T$, the goal is to produce a logical decision $D$. 
When applying MLLM to a VLC problem, it can be formulated as $\hat{D} = \mathcal{F}_{MLLM}(I,T)$, where $\hat{D}$ denotes the predicted decision and $\mathcal{F}_{MLLM}()$ represents the function of MLLM to predict the correct decision. Recent studies show that an MLLM functions as a statistical prediction function, which predicts the next token based on the preceding sequence of visual and textual tokens, according to learned probability distributions. Hence, the function of the MLLM can be expressed as 
\begin{equation}\label{eq:mllm-function}
\hat{D} = \mathcal{F}_{MLLM}(I,T) = \arg \max\nolimits_{D \in \mathcal{D}} P_M(D|I,T),
\end{equation}
where $\mathcal{D}$ represents a potential decision set. As MLLM makes prediction directly from the visual language inputs without explicit reasoning, it acts as a black-box. This often leads to factual inaccuracies~\cite{awais2023amber} and hallucinations~\cite{Huang_2025, Rahman_2026, bai2025hallucinationmultimodallargelanguage}, especially when the model is applied to novel tasks without gt annotations. 

In contrast, humans justify decisions using explicit visual evidence and concrete logic rules~\cite{10.3389/frai.2022.806403}. Various modern foundation models can provide complementary strengths that support such explicit reasoning. LLMs excel at language understanding and semantic reasoning~\cite{zhao2025surveylargelanguagemodels}, MLLMs extend this capability to vision-language descriptions and VQA~\cite{liu2023llava,Chen_2024_CVPR_InternVL,bai2023qwenvl}, and VFMs are pre-trained for fundamental visual tasks such as classification, segmentation and detection~\cite{shen2024groundvlp,kirillov2023segany}. Although these models are trained with different annotations, they share large overlapping public datasets. Therefore, they should provide consistent and complementary information when given the same same multimodal input. Hence, we propose a Logic Channel, that operates in parallel with MLLM, to support model validation, selection, justification and enhancement without requiring ground-truth. Given a VLC problem, the logic channel can be formulated as a three-step operation. First, a set of task-relevant facts and logical relations is extracted from the text expression by prompting an LLM. This can be expressed as $(\hat{F}_s, \hat{R}_s) = \mathcal{F}_{LLM}(T|\text{prompt})$, where the subscript $s$ indicate the semantic-level representation. Next, a VFM is employed to ground each extracted fact in the query image $I$, and produce corresponding confidence probabilities, \ie, $\hat{F}_v = \mathcal{F}_{VFM}(I|\hat{F}_s)$, where the subscript $v$ denotes the grounded visual evidence. Finally, basic logical rules for factual, counter-factual, and relational reasoning are applied to derive the logical inference as
\begin{equation}\label{eq:elc-function}
\hat{D}_L = \mathcal{F}_{LR}(D|I,T) = \arg \max\nolimits_{D \in \mathcal{D}} P_{LR}(D|\hat{F}_v,\hat{R}_s),
\end{equation}
where the subscript $LR$ represents Logic Reasoning.

The proposed dual-channel framework is illustrated in Figure~\ref{fig:dual-channel-framework}. The upper channel employs an MLLM to produce prediction directly as a black box, forming the {\em Implicit Logic Channel} (ILC). The lower channel, mimicking human logical reasoning, employs an LLM and VFM to explicitly extract and ground facts, followed by applying logical reasoning on the grounded facts for final decision, forming the {\em Explicit Logic Channel} (ELC). When the foundation models understand the VLC problem well, the two channels should be consistent and complementary.
The logical consistency between the two channels provides principled basis for model validation, selection, justification and enhancement in zero-shot applications.\\
\noindent {\bf Validation}: 
When applying an MLLM to a new VLC task, it is common to have a set of test samples without gt annotations. In such cases, the proposed dual-channel system allows us to evaluate the logical consistency between the ILC and ELC. Let $\mathcal{Q}$ represents the test set and $q \in \mathcal{Q}$ is a test sample. We define the Consistency Rate ($CR$) between ILC and ELC as
\begin{equation}\label{eq:cr}
    CR=\frac{1}{|\mathcal{Q}|} \sum\nolimits_{q \in \mathcal{Q} } \mathbb{I}(\hat{D}(q) = \hat{D}_L(q)),
\end{equation}
where the indicator function $\mathbb{I}$ returns 1 if the two channels produce consistent predictions and 0 otherwise. For different VLC task, a suitable indicator function has to be implemented. A higher $CR$ indicates that the MLLM is more reliable and logical for the new task. In practice, $CR$ reflects three general situations:
\begin{itemize}
\item Well-covered knowledge. If the knowledge required for the VLC task is well represented in the shared training datasets of the foundation models, both channels tend to produce correct predictions, resulting in high $CR$. 
\item Partially covered knowledge. If only partial knowledge is presented in the shared training data, one of the two channels may fail to perform correctly, leading to lower $CR$. 
\item Novel or OOD scenarios. If both the language expressions and visual content are novel compared to the existing training datasets, e.g., out-of-distribution (OOD) scenarios, both channels may fail, and the $CR$ score becomes low.  
\end{itemize}

Therefore, $CR$ can be used as a principled metric for zero-shot model selection when the gt annotations are not available for a new VLC task. The discrepancies between the two channels also provide cues to user to perform manual validation and diagnose the situation without the cost of gt annotations.\\
\noindent {\bf Enhancement}: The two channels provide complementary strengths, and fusing their outputs can further improve prediction accuracy. The logical consistency serves as a guide for aligned fusion. 
In principle, when the ILC and ELC predictions are logically consistent, the prediction is highly likely to be correct. Our experiment results strongly support this assumption. Let $\mathcal{Q}_c \in \mathcal{Q}$ be the consistency subset, where, for each $q \in \mathcal{Q}_c$, $\mathbb{I}(\hat{D}(q)=\hat{D}_{LR}(q))=1$, indicating that $\hat{D}(q)$ and $\hat{D}_{LR}(q)$ are reliable predictions from the ILC and ELC channels, respectively. 
For all samples in the consistent subset $\mathcal{Q}_c$, we compute the mean confidence scores of the reliable predictions for both channels as
\begin{equation}
    \mu_{ILC}^c = \frac{1}{| \mathcal{Q}_c |} \sum\nolimits_{q \in \mathcal{Q}_c} P_M(D|q) \quad \textrm{and} \quad  \mu_{ELC}^c = \frac{1}{| \mathcal{Q}_c |} \sum\nolimits_{q \in \mathcal{Q}_c} P_{LR}(D|q).
\end{equation}
Without gt labels, for $q \in \mathcal{Q}_c$, predictions from both channels should be trusted equally. When applied to a new test example $q_n$, the probability of aligned fusion can be computed as
\begin{equation}\label{eq:fusion_mca}
    P_F(D|q_n) = P_M(D|q_n) + \mu_{ILC}^c [P_{LR}(D|q_n)/\mu_{ELC}^c],
\end{equation}
where the prediction from ELC is first normalized on $\mu_{ELC}^c$ and re-scaled to align with ILC. Aligned scores are used for final decision on maximum. 
Beyond prediction, the difference between the final decision and predictions from ILC and ELC would provide additional information on the reliabilities of the channels.

\subsection{Logic Consistency for MC-VQA on Factual Evidence}

\begin{figure}[tb]
  \centering
  \includegraphics[width=0.9\columnwidth]{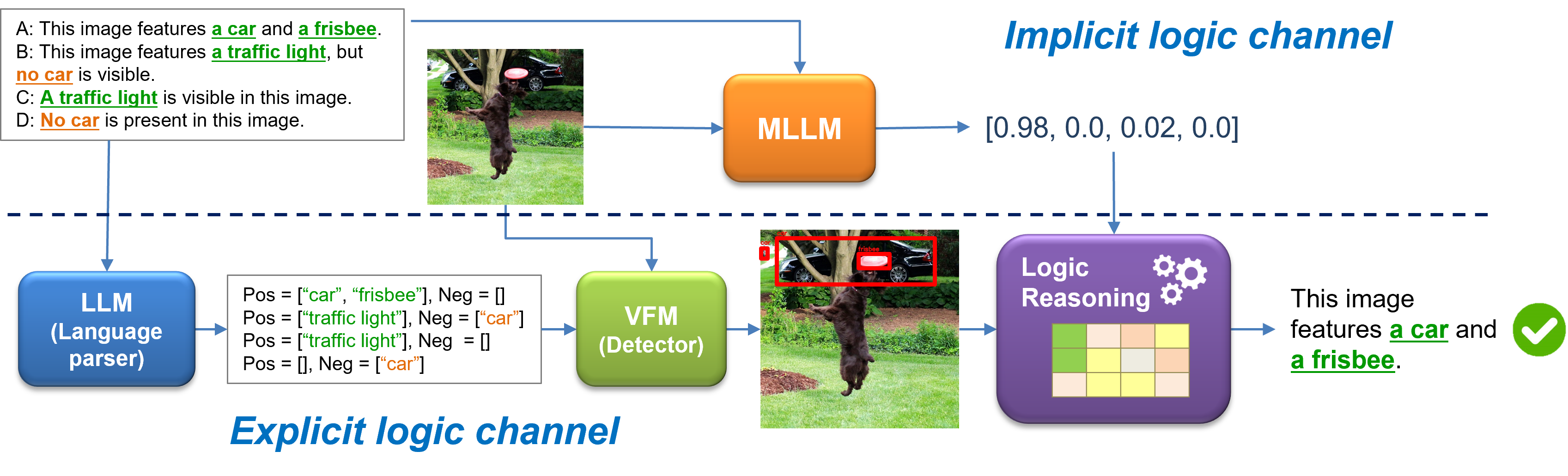}
  \caption{ELC (Explicit Logic Channel) and logic consistency for MC-VQA task on factual and counter-factual reasoning, with an example from NegBench for illustration.}
  \label{fig:framework_negbench}
\end{figure}

A representative task of VLC is multiple-choice VQA (MC-VQA). Given an image $I$, a question $T$ and a set of candidate answers $\mathcal{A}=[a_1, \cdots, a_K]$, the model must choose the correct answer ($\hat{c} \in [1,K]$). A related variant replaces the question with a set of candidate captions $\mathcal{T} = [T_1, \cdots, T_K]$, where the model selects the most suitable caption ($\hat{c} \in [1,K]$). 
It can be formulated as
\begin{equation}
    \hat{c} = \arg \max_{c \in [1,K]} P_M(c|I,T,\mathcal{A}) \quad \textrm{or} \quad \hat{c} = \arg \max_{c \in [1,K]} P_M(c|I,\mathcal{T}).
\end{equation}
In both cases, the MLLM operates as a black-box choice function.

To evaluate joint vision-language understanding, MC-VQA datasets typically design questions or captions that reference objects or concepts that are present or absent from the image. Logically, solving such tasks requires decisions grounded in factual evidence (presence of positive elements) and counter-factual (absence of negative elements). As modern LLMs possess powerful capabilities in language understanding, concept extraction, and semantic reasoning, for each text query, we prompt the LLM to extract positive (pos) and negative (neg) objects or concepts for explicit logical justification. 

The dual-channel system for MC-VQA is presented in Figure~\ref{fig:framework_negbench}, with an example from NegBench for illustration. 
For each question, the MLLM receives the image, the text query, and answer choices, and is prompted to output both the predicted answer and confidence values (0\%-100\%) for all $K$ options (\ie, $P_M()$). In the ELC, an LLM is first prompted to extract the {\em pos} and {\em neg} nouns from the text query. These nouns are then passed to a VFM. For each object category, the VFM locates all instances in the image and computes their detection probabilities. The probability of an object category is obtained as the maximum over all its detected instances. Let there be $K$ pos object categories and $L$ neg object categories, their presence probabilities are denoted as $\{ P(O_p^k) \}_{k=1}^K$ and $\{ P(O_n^l) \}_{l=1}^L$. The presence of pos objects provides factual evidence on $T$, while the presence of neg objects provides counter-factual evidence. The corresponding probabilities are computed logically as
\begin{equation}\label{eq:pos_prob}
    P(pos)=\min \{ P(O_p^1), \cdots , P(O_p^K)  \}, ~ P(neg) = \max \{ P(O_n^1), \cdots, P(O_n^L) \}.
\end{equation}
The final probability of factual and counter-factual evidence is computed as
\begin{equation}\label{eq:probvt}
    P_{LR}(c|I,T) = \left\{
    \begin{array}{l}
    P(pos), \quad \quad neg=\emptyset \\
    1-P(neg), \quad pos = \emptyset \\
    \left[ P(pos) (1-P(neg)) \right]^{\frac{1}{2}}, ~ pos \ne \emptyset ~\&~ neg \ne \emptyset
    \end{array}
    \right.
\end{equation}
where the geometric mean is used for normalization as done in~\cite{malinin2020uncertainty}. The ELC selects the choice with the maximum posterior $P_{LR}(T|I)$ as the correct answer. 

For validation, the two channels are run independently, and $CR$ is then computed. With a small set of logical consistent samples, the system can further enhance the performance by applying the aligned fusion strategy in Eq.~(\ref{eq:fusion_mca}), fusing outputs from both the ILC and ELC channels.

\subsection{Logical Consistency for HC-REC on Association}
\label{subsection:hc-rec}

Referring Expression Comprehension (REC) is another representative task in VLC. Given an image $I$ and a text expression $T$, the goal is to localize the referred object ($O$). Recent Human-Centric REC (HC-REC) benchmarks introduce rich or long contextual descriptions, making them challenging for VLC~\cite{NEURIPS2024_80f0cd03}. In HC-REC, correct prediction depends on the presence of the target person, associated objects, and understanding their visual relations in the image. 

\begin{figure}[tb]
  \centering
  \includegraphics[width=0.9\columnwidth]{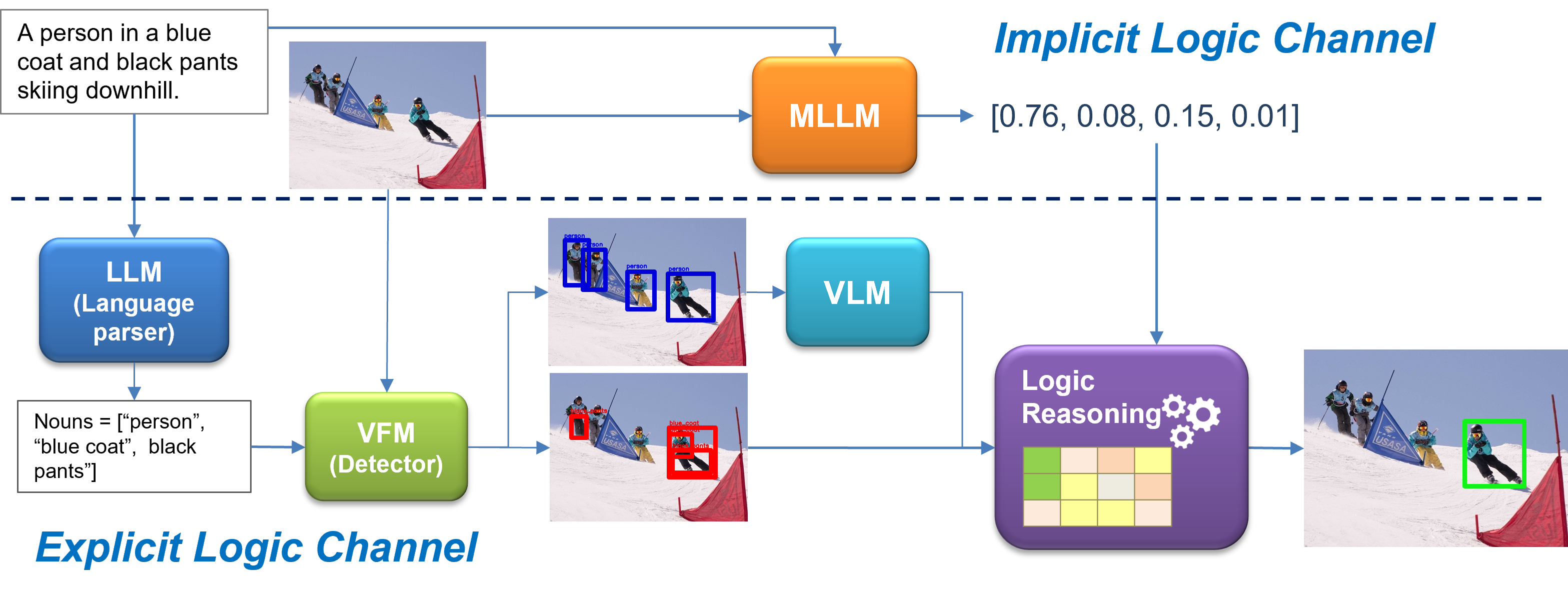}
  \caption{ELC (Explicit Logic Channel) and logic consistency for HC-REC task on object association, with an example from HC-RefCOCOg for illustration.}
  \label{fig:framework_refcocog}
\end{figure}

The dual-channel system for HC-REC is shown in Figure~\ref{fig:framework_refcocog}, with an example from HC-RefCOCOg for illustration. In ILC, the MLLM receives the image and text query, and is prompted to directly predict the bounding box coordinates of the referred person, along with its confidence score (\ie, $P_M(h_m|T)$). In ELC, an LLM is prompted to extract person-related and object-related nouns from $T$. Then, these nouns are passed to a VFM to locate all instances of the described persons and associated objects. Each detected person is cropped, and fed to a VLM along with the referring expression to obtain a vision-language matching score. For associated objects, their presence probabilities are based on detection confidence scores. Assume that there are ($K+1$) extracted nouns, which consist of one person $H$ and $K$ object concepts, denoted as $\{ H, O_1, \cdots, O_K \}$. The VFM locates a set of instances for person and objects. 
For each detected instance, let $h_i$ denote the $i$th detected person, and $o_{jl}$ the $l$th instance of object $O_j$.

To estimate the probability of each person being the target, we adopt an evidence-accumulation approach, as inspired by the principle of perceptual decision-making~\cite{Balsdon:20}. Formally, it can be expressed as
\begin{equation}\label{eq:evi_relations}
    P_{LR}(h_i|T) = \frac{1}{K+1} \left( \sum\nolimits_{k=1}^K P(O_k|h_i) + P(h_i|H) \right),
\end{equation}
where $P(h_i|H)$ represents the presence probability of person $h_i$, and $P(O_k|h_i)$ denotes its association with object $O_k$. Assuming that there are $L$ detected instances of object $O_k$, i.e., $\{ o_{kl} \}_{l=1}^L$, the association rate between $h_i$ and $o_{kl}$ can be computed as
\begin{equation}\label{eq:association_rate}
    R_A(o_{kl},h_i) = A_{int}(o_{kl}, h_i) / A(o_{kl}),
\end{equation}
where $A_{int}(o_{kl}, h_i)$ denotes the intersection area of $o_{kl}$ and $h_i$, and $A(o_{kl})$ is the area of $o_{kl}$. The overall association probability between $h_i$ and $O_k$ becomes
\begin{equation}\label{eq:evidence_association}
    P(O_k|h_i) = \max\nolimits_{l \in [1,L]} \left\{ R_A(o_{kl}, h_i) \right\}.
\end{equation}
Finally, the person with the highest accumulated probability $P_{LR}(h_i|T)$ is selected as the grounded person in the ELC.

For model validation, both channels operate independently and the $CR$ is computed across multiple IoU levels, \eg, IoU thresholds of 0.5, 0.75, and 0.9. When fusing two channels for performance enhancement, we incorporate IoU-weighted probabilities for each candidate person, and extend the aligned fusion Eq. ~(\ref{eq:fusion_mca}) to HC-REC as
\begin{equation}\label{eq:fusion_ref}
    \left\{
    \begin{array}{lll}
    P_F(h_m|T) & = & P_M(h_m|T) + \max_{i \in [1,K]} [IoU_{mi} \cdot (\frac{\mu_{ILC}^c}{\mu_{ELC}^c} P_{LR}(h_i|T))] \\
    P_F(h_i|T) & = & IoU_{im} \cdot P_M(h_m|T) + \frac{\mu_{ILC}^c}{\mu_{ELC}^c} P_{LR}(h_i|T)
    \end{array}
    \right.
\end{equation}
The person with the highest $P_F()$ value is selected as the final prediction.

\subsection{Logic Consistency for HC-REC on Correlation}

Recent VLC tasks involving very long referring expressions have introduced new challenges for MLLM, especially in zero-shot settings. With the powerful capabilities of LLMs in language understanding, summarization and semantic categorization, we can leverage these strengths by prompting the LLM to extract essential visual facts from long text descriptions, and use them to guide explicit logical reasoning (\ie, ELC) for visual-language verification. 

\begin{figure}[tb]
  \centering
  \includegraphics[width=1.0\columnwidth]{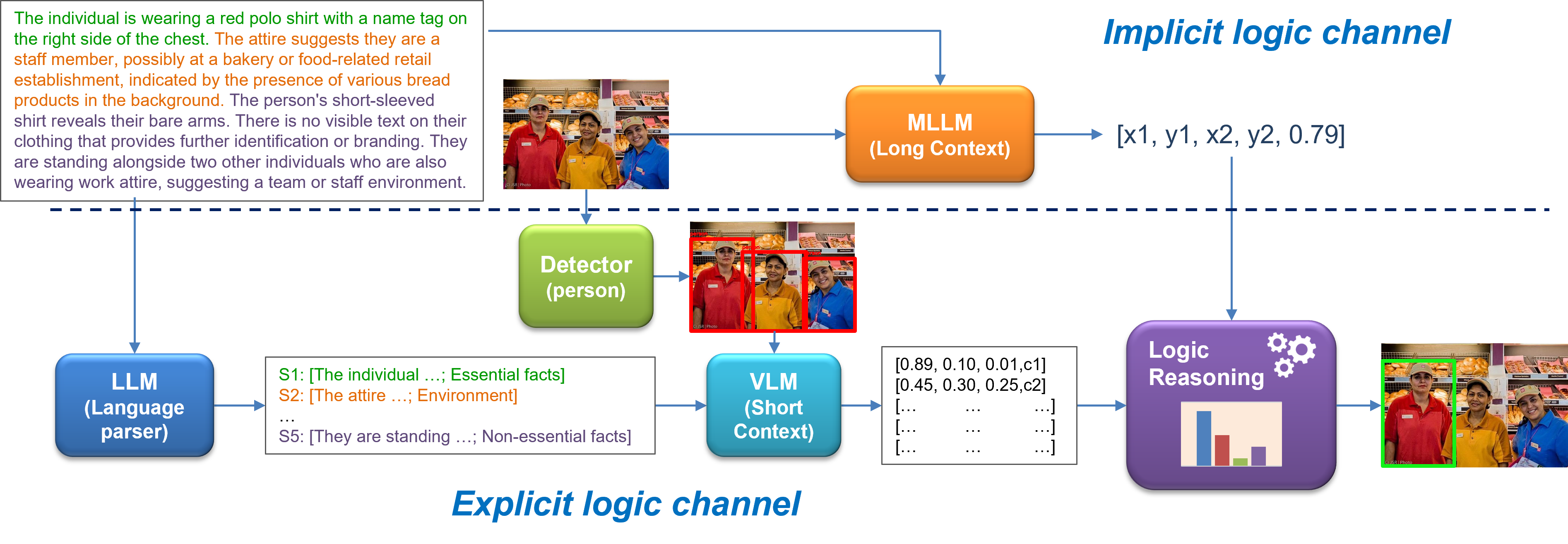}
  \caption{ELC (Explicit Logic Channel) and logic consistency for task of HC-REC on long context text query, with an example from HC-RefLoCo for illustration.}
  \label{fig:framework_refloco}
\end{figure}

In HC-REC with very long context input, the challenge is not only the text length, but also the presence of many non-informative or weakly relevant sentences. For VLC, sentences containing essential visual facts are far more informative for locating the target person than sentences describing general context or unrelated events~\cite{Cheng2025TheFL}. Consider the following example of long referring expression for an image of human activities in a public park: ``{\em This is a scene of a public park in Sunday. People play various activities in the morning. A young lady in red sportswear is sitting on a bench for a rest. A blue bottle is placed on the bench beside her.}'' The first sentence ($S_1$) is a description of the environment, and does not lead the attention to the target young lady ($H$). The second ($S_2$) is not closely related and may incorrectly draw attention to every person in $I$. The third ($S_3$) provides the core description that uniquely identifies the target $H$. The fourth ($S_4$) provides secondary but relevant information. Following the recommendations in~\cite{Cheng2025TheFL}, we prompt the LLM to classify each sentence into one of the three categories, \ie, Essential Fact, Non-Essential Fact or Environment, for the task of HC-REC. The full prompt template is presented in Suppl. Mat.

The dual-channel system for HC-REC with long context expression is presented in Figure~\ref{fig:framework_refloco}, with an example from HC-RefLoCo for illustration. The ILC operates the same as in Figure~\ref{fig:framework_refcocog}, while the ELC is adapted to focus on the effectiveness of sentence-level information. The long context expression is first split into sentences and the LLM is prompted to classify each sentence into one of the three categories. Then, a VFM detects all persons in the image and each person box is cropped. Each cropped person patch is paired with each sentence to form a VL pair, and fed to a VLM to obtain a VL matching probability.
If there are $N$ detected persons and $K$ sentences, the set of predicted Human-Sentence matches can be denoted as $\{\{ P_{VLM}(h_n|S_k) \}_{k=1}^K \}_{n=1}^N$. Naturally, pairings between Essential Facts and the correct person would results in higher probabilities, while other pairs produce lower matching scores. Following the principle of human decision-making theory~\cite{Bradley2012-BRADTA-6}, the final logical prediction can obtained as a weighted function
\begin{equation}\label{eq:L-decision-making}
    P_{LR}(h_n|T) = \sum\nolimits_{k=1}^K P_{VLM}(h_n|S_k) u(S_k)
\end{equation}
where $u(S_k)$ is a utility weight of $S_k$ that reflects its informativeness for perceptual decision-making. On common-sense knowledge, the sentences of Essential Facts should receive higher utility weight, Non-Essential Facts sentences should have moderate weight, and the Environment sentences should receive very low weight. This common-sense weighting is effective for perceptual decision-making especially in zero-shot application scenarios. The person with the highest accumulated score is selected as logical prediction of ELC.

Model validation and enhancement follow the same procedure described in previous subsection~\ref{subsection:hc-rec} on HC-REC task.

\section{Experiments}
\label{sec:exp}
We conduct comprehensive experiments to evaluate the effectiveness of ELC and CR for model validation and enhancement when applying MLLMs to novel VLC tasks under zero‑shot settings without gt annotations. Evaluations are performed across two representative VLC tasks on three recent challenging benchmarks, with 11 frontier open‑source MLLMs from four model families.

\noindent {\bf Tasks and Datasets}: \\
{NegBench~\cite{alhamoud2025vision} is a recent benchmark designed to assess visual–language understanding under factual and counter‑factual reasoning. Each sample contains both positive and negative phrases, and choices follow three linguistic templates: Affirmation, Negation, and Hybrid. An affirmation text contains only positive elements $\{pos\}$, i.e., objects present in the image. A negation text includes only negative elements $\{ neg \}$, i.e., objects absent from the image but commonly associated with the present objects. A hybrid text contains both positive and negative elements. There are three natural image MCQ tasks, \ie, COCO, VOC2007, and HardNeg-Syn. Our experiments are conducted on the publicly available data, i.e., 5914 MCQ questions on COCO and 5032 MCQ questions on VOC2007.

Recent HC-REC (Human-Centric Referring Expression Comprehension) benchmarks, extended from general REC task, are created for evaluating VLC capabilities in MLLMs~\cite{NEURIPS2024_80f0cd03}. RefCOCOg~\cite{mao2016refcocog} is created from COCO with enriched phrases (average 8.9 words), as compared to RefCOCO/+ (3.3 and 3.4 words), providing a more challenging scenario.

HC-RefLoCo (Human-Centric Referring Expression Comprehension with Long Context)~\cite{NEURIPS2024_80f0cd03} is a recent challenging benchmark featuring long-context expressions with an average of 93.2 words. 
It is built upon various image datasets, including COCO, Objects365, OpenImage v7, and LAION-5B, covering a wide range of real-world scenes. These datasets allow us to examine the robustness of MLLMs under both moderate-length and extremely long referring expressions.

\noindent {\bf Experimental settings}: 
ILC predictions are generated using the default prompt templates provided by each benchmark. In ELC, we employ Qwen3-4B-Instruct~\cite{qwen3technicalreport} as the LLM, EvaCLIP-8B~\cite{EVA-CLIP-18B} as the VLM for HC-RefCOCOg and InternVL2.0-8B~\cite{Chen_2024_CVPR_InternVL} as the VLM for HC-RefLoCo. All models are moderate-sized, with no fine-tuning or additional optimization is applied. Extensive ablation evaluations on ELC, including reliability on model variations, critical false positives, confusion matrix, and latency analysis, can be found in the Suppl.

\noindent {\bf Experimental objectives}:\\ 
Our experiments are designed to answer three key questions: 
(1) Is CR score meaningful without gt. To assess whether CR can serve as a reliable metric in zero‑shot evaluation, we compute CR between ILC and ELC, and compute accuracy (Acc) based on gt for analysis. We measure their relationship using three representative correlation coefficients~\cite{AKOGLU201891}: Pearson's r ($r$), Spearman's rho ($\rho$), and Kendall's tau ($\tau$). $r$ compares the scores of the metrics, while $\rho$ and $\tau$ compare the rankings on the metrics. Each coefficient ranges from -1 to 1, where values near 1 indicate strong positive correlation, values between 0.5 and -0.5 indicate weak or no correlation, and values near -1 indicate strong negative correlation. (2) Is CR score helpful for model validation and selection without gt. On each of the three benchmarks, we evaluate 11 frontier MLLMs and analyze their performance, e.g., reliable predictions, strong models. (3) Effectiveness of ELC and CR in enhancing MLLM performance for VLC without gt. We evaluate the effect of aligned fusion between ILC and ELC, and examine the performance gain across benchmarks. Our results support that ELC and CR provide consistent performance improvements without requiring any additional training, confirming its practicality for zero-shot applications. 

\subsection{Experimental Results}

\noindent{\bf MC-VQA Results on NegBench}: 
Table~\ref{tab:negbench} summarizes the performance of the 11 MLLMs on NegBench. The column $Acc$/Rank reports the Accuracy score and its corresponding ranking. $CR$/Rank presents the CR score and its ranking, and $Acc_E$ shows the Accuracy of enhancement after applying aligned fusion. We observe that both InternVL and Qwen are strong model families, where three out of four models achieve high $Acc$/Rank and $CR$/Rank across COCO and VOC2007 datasets. 

\begin{table}[th]
  \caption{Results on NegBench for MC-VQA task.}
  \label{tab:negbench}
  \centering
  \resizebox{0.8\textwidth}{!}{
  \begin{tabular}{l|l|l|l|l|l|l}
    \toprule
     NegBench & \multicolumn{3}{c|}{COCO} & \multicolumn{3}{c}{VOC2007} \\ 
    \midrule    
    MLLM & $Acc$/Rank & $CR$/Rank & $Acc_E$ & $Acc$/Rank & $CR$/Rank & $Acc_E$ \\
    \midrule
    Gemma-3-12B \cite{gemma_2025}  & 0.7198/6 & 0.6715/6 & 0.7201 & 0.8750/5 & 0.8006/5 & 0.8752 \\
    InternVL2.0-8B\cite{Chen_2024_CVPR_InternVL}  & 0.4878/9 & 0.3980/10 & 0.8434 & 0.5868/10 & 0.5198/10 & 0.9352 \\
    InternVL2.5-8B\cite{Chen_2024_CVPR_InternVL}  & 0.9121/2 & 0.7038/4 & {\bf 0.9650} & 0.9201/3 & 0.8076/4 & {\bf 0.9809}  \\
    InternVL3.0-8B\cite{Chen_2024_CVPR_InternVL}  & {\bf 0.9319}/1 & {\bf 0.7259}/1 & 0.9557 & {\bf 0.9537}/1 & {\bf 0.8424}/1 & 0.9684 \\
    InternVL3.5-8B\cite{Chen_2024_CVPR_InternVL}  & 0.8429/3 & 0.6583/7 & 0.9324 & 0.9221/2 & 0.8088/3 & 0.9750  \\
    LLaVA-1.5-13B\cite{liu2023llava}  & 0.6206/8 & 0.5810/8 & 0.6206 & 0.7738/8 & 0.7112/8 & 0.7738 \\
    LLaVA-1.6-13B\cite{liu2023llava}  & 0.3898/10 & 0.4004/9 & 0.4499 & 0.6528/9 & 0.6096/9 & 0.6897  \\
    QwenVL-7B-Chat\cite{bai2023qwenvl}  & 0.2426/11 & 0.2310/11 & 0.6245  & 0.2371/11 & 0.2314/11 & 0.5560 \\
    Qwen2.0-VL-7B\cite{bai2023qwenvl}  & 0.6968/7 & 0.6743/5 & 0.7076 & 0.8535/6 & 0.7822/6 & 0.8764 \\
    Qwen2.5-VL-7B\cite{bai2023qwenvl} & 0.8165/4 & 0.7109/3 & 0.8356 & 0.9197/4 & 0.8374/2 & 0.9223  \\
    Qwen3.0-VL-8B\cite{bai2023qwenvl}  & 0.7753/5 & 0.7134/2 & 0.7758 & 0.8408/7 & 0.7688/7 & 0.8408  \\
    \midrule
    ELC  & 0.7577 &  &  & 0.8684 & &  \\
  \bottomrule
  \end{tabular}
  }
\end{table}

\noindent{\bf HC-REC Results on HC-RefCOCOg and HC-RefLoCo}:\\
The results of 11 MLLMs on HC-RefCOCOg are presented in Table~\ref{tab:HC-RefCOCOg}. The column $Acc$/50/75/90/R denote Accuracy scores obtained at IoU thresholds 0.5, 0.75 and 0.9, and the ranking based on Acc50. Corresponding $CR$ is also computed at IoU thresholds 0.5, 0.75, and 0.90. CR50 scores are presented here (see full results in Suppl. Mat.). The column $CR$/R denotes CR50 scores and its ranking. The enhanced performance is evaluated on test set with the alignment weight obtained on val set. The Accuracy scores of enhanced performance are added in column $A_E$/50/75/90 of test set. Table~\ref{tab:HC-RefLoCo} presents the results of 11 MLLMs on HC-RefLoCo. Here, the column $mAcc$ denotes the mean Accuracy over IoU from 0.1 to 0.9 with step size 0.1. Again, enhanced performance is evaluated on test set. Only $Acc$50 and $mAcc$50 are presented (see Suppl. Mat.).

\begin{table}[th]
  \caption{Results on HC-RefCOCOg for HC-REC task.}
  \label{tab:HC-RefCOCOg}
  \centering
  \resizebox{0.8\textwidth}{!}{
  \begin{tabular}{l|l|l|l|l|l}
    \toprule
     HC-RefCOCOg & \multicolumn{2}{c|}{val} & \multicolumn{3}{c}{test} \\ 
    \midrule     
    MLLM & $Acc$/50/75/90/R & $CR$/R & $Acc$/50/75/90/R & $CR$/R & $A_E$/50/75/90 \\
    \midrule
    Gemma-3-12B  & .088/.004/.001/10 & .081/11 & .088/.004/.001/10 & .082/11 & .752/.690/.587 \\
    InternVL2.0-8B  & .751/.657/.386/4 & .665/4 & .755/.672/.408/4 & .663/3 & .841/.777/.637 \\
    InternVL2.5-8B  & .765/.710/.517/3 & .670/3 & .775/.722/.533/3 & .657/4 & .850/.791/.659  \\
    InternVL3.0-8B  & .699/.620/.403/6 & .618/6 & .713/.639/.427/6 & .613/7 & .837/.771/.633  \\
    InternVL3.5-8B  & .780/.735/.597/2 & .681/2 & .808/.761/.610/2 & .681/2 & .854/.796/.670  \\
    LLaVA-1.5-13B  & .219/.069/.012/9 & .224/9 & .236/.076/.015/9 & .231/9 & .768/.695/.586  \\
    LLaVA-1.6-13B  & .076/.007/.001/11 & .089/10 & .082/.005/.002/11 & .087/10 & .752/.693/.591 \\
    QwenVL-7B-Chat  & .391/.361/.300/8 & .411/8 & .391/.365/.305/8 & .395/8 & .780/.724/.616 \\
    Qwen2.0-VL-7B  & .715/.631/.489/5 & .629/5 & .735/.647/.507/5 & .629/5 & .833/.769/.643 \\
    Qwen2.5-VL-7B  & .684/.591/.395/7 & .612/7 & .709/.620/.390/7 & .618/6 & .838/.776/.642  \\
    Qwen3.0-VL-8B  & .804/.749/.597/1 & .703/1 & .818/.777/.632/1 & .692/1 & {\bf .856/.798/.673} \\
    \midrule
    ELC  & .807/.747/.628 & & .801/.743/.634 & & \\   
  \bottomrule
  \end{tabular}
  }
\end{table}

\begin{table}[th]
  \caption{Results on HC-RefLoCo for HC-REC task.}
  \label{tab:HC-RefLoCo}
  \centering
  \resizebox{0.8\textwidth}{!}{
  \begin{tabular}{l|l|l|l|l|l|l|l|l}
    \toprule
     HC-RefLoCo & \multicolumn{3}{c|}{val} & \multicolumn{5}{c}{test} \\ 
    \midrule     
    MLLM & $Acc$50 & $mAcc$/R & $CR$/R & $Acc$50 & $mAcc$/R & $CR$/R & $Acc_E$50 & $mAcc_E$   \\
    \midrule
    Gemma-3-12B  & .0921 & .0219/10 & .0860/10 & .0920 & .0220/10 & .0860/10 & .4270 & .3500  \\
    InternVL2.0-8B  & .7592 & .5403/4 & .6802/4 & .7534 & .5350/4 & .6860/4 & .8120 & .6412 \\    
    InternVL2.5-8B  & .5039 & .3048/7 & .4447/7 & .4970 & .3041/7 & .4478/7 & .6618 & .5144 \\
    InternVL3.0-8B  & .6817 & .4790/6 & .6100/6 & .6742 & .4720/6 & .6095/6 & .7630 & .6071 \\
    InternVL3.5-8B  & .8832 & .6718/2 & .7691/2 & .8866 & .6733/2 & .7746/2 & .9003 & .7323  \\
    LLaVA-1.5-13B  & .1906 & .0821/9 & .1936/9 & .1957 & .0834/9 & .1966/9 & .4304 & .3353  \\
    LLaVA-1.6-13B  & .0626 & .0140/11 & .0612/11 & .0621 & .0144/11 & .0589/11 & .3335 & .2797 \\
    QwenVL-7B-Chat  & .3284 & .2687/8 & .3313/8 & .3314 & .2699/8 & .3332/8 & .4959 & .4303 \\
    Qwen2.0-VL-7B  & .8396 & .6532/3 & .7375/3 & .8451 & .6542/3 & .7449/3 & .8652 & .7123 \\
    Qwen2.5-VL-7B  & .7502 & .5283/5 & .6604/5 & .7522 & .5288/5 & .6650/5 & .8260 & .6526  \\
    Qwen3.0-VL-8B  & .9412 & .7992/1 & .8045/1 & .9418 & .8024/1 & .8128/1 & .9388 & {\bf .8143}  \\
    \midrule
    ELC  & .804 & .723 & & .815 & .734 &  &  &  \\    
  \bottomrule
  \end{tabular}
  }
\end{table}

\noindent{\bf Correlation Analysis}: Based on the accuracy and CR scores across the three benchmark tables, we compute the correlation coefficients between Acc and CR for all models and presented in Table~\ref{tab:correlation}.

\begin{table}[th]
  \caption{Correlation coefficients between CR and Acc.}
  \label{tab:correlation}
  \centering
  \resizebox{0.7\textwidth}{!}{  
  \begin{tabular}{l|c|c|c|c|c|c|c}
    \toprule
    Benchmark & \multicolumn{2}{c|}{NegBench} & \multicolumn{2}{c|}{HC-RefCOCOg} & \multicolumn{2}{c|}{HC-RefLoCo} & All \\ 
    \midrule    
    Cor. Coefficient & COCO & VOC2007 & val & test & val & test & mean \\
    \midrule
    Pearson's $r$ & 0.9543 & 0.9968 & 0.9972 & 0.9973 & 0.9987 & 0.9987 & 0.9905 \\
    Spearman's $\rho$ & 0.8364 & 0.9727 & 0.9909 & 0.9727 & 1.0000 & 1.0000 & 0.9621\\
    Kendall's $\tau$ & 0.6727 & 0.9273 & 0.9636 & 0.8909 & 1.0000 & 1.0000 & 0.9091 \\   
    \bottomrule
  \end{tabular}
  }
\end{table}

\noindent{\bf Latency Analysis}:
Naturally, ELC introduces additional time cost. The main time costs are spent by LLM for language parser, VFM for human and object grounding, and VLM for vision-language association. The ELC time cost information on the three benchmarks are presented in Table~\ref{tab:ELC-time-cost}.
\begin{table}[th]
  \caption{Latency analysis on ELC on the tasks of the three benchmarks.}
  \label{tab:ELC-time-cost}
  \centering
  \resizebox{0.9\textwidth}{!}{
  \begin{tabular}{l|l|l|l}
    \toprule
    Benchmark & LLM (Language Parser) & VFM (Object Grounding) & VLM (Visual-Text Association) \\
    \midrule
    NegBench & Mistral: 1.80s & GroundingDINO: 0.76s & - \\    
    HC-RefCOCOg & Mistral: 1.77s & GroundingDINO: 0.46s & EvaCLIP: 1.45s \\
    HC-RefLoCo & Mistral: 1.55s & GroundingDINO: 0.35s & EvaCLIP: 6.88s \\
  \bottomrule
  \end{tabular}
  }
\end{table}

\subsection{Evaluation and Discussion}

\noindent {\bf Effectiveness of CR}:
As shown in Table~\ref{tab:correlation}, the correlation between CR and Acc is consistently strong across all benchmarks. All the correlation measures,  $r$, $\rho$ and $\tau$, exceed 0.89, except on COCO of NegBench where $\rho$>0.83 and $\tau$>0.67, still well above 0.5. The average scores of the three metrics exceed 0.9, indicating very strong correlation between CR with Acc. These results confirm thatCR is a reliable indicator of model performance even without gt annotation. Consistent but Incorrect (CI) errors might cause risks, especially when CR is high. We evaluated such cases. As example, on HC-RefCOCOg with Qwen3.0-VL, CR is 0.69 (top 1), the rate of CI is 4.7\%, and manual examination found 52\% caused by inconsistency with gt boxes, 32\% related to bias of models, and only 15\% caused by ELC errors. Once CR is high, CI risk is low. Full results and visual examples can be found in Suppl. Sec 2.2.

\noindent {\bf Effectiveness for Model Validation and Selection}: From the columns of CR and Acc, we observe substantial performance variation among recent frontier MLLMs. For example, CR on NegBench COCO set ranges widely from 0.23 to 0.73. Even within the InternVL family, CR varies from 0.40 to 0.73. Besides, newer version models are not guaranteed to outperform earlier ones, even from the same family, \eg, InternVL-3.0 is better than InternVL-3.5 on NegBench. In additional, a model that excels on one task may perform poorly on another, \eg Gemma-3 achieves strong results on MC-VQA but fails on HC-REC. These findings highlight that model selection is challenging in zero-shot scenarios due to inconsistent model performance across tasks. The proposed ELC and CR provide a principled mechanism to validate and select suitable models without requiring annotated datasets.  
In addition, the inconsistent samples would lead to efficient manual examination and reveal informative insights (see Suppl.).

\noindent {\bf Effectiveness for Performance Enhancement}: 
To validate the aligned fusion assumption, we compute $CR_{gt}$ on samples that are both consistent and correct when given ground-truth. The ratio $CR_{gt}/CR$ is computed for every model across all tasks. Even when $CR$ varies widely from 0.1 to 0.95, the ratio remains above 80\%. The mean of ratio is 0.92 at $CR$=0.5508, and increases linearly with $CR$, indicating that higher consistency leads to higher correctness. However, even the rate is low, the cases of consistent but incorrect (not matching gt label) might raise concerns on the validation of CR. Additional investigation is performed (see Suppl. Sec 2.2). 

Aligned fusion yields consistent improvements over MLLMs across all three benchmarks. Even the strongest model benefits from ELC-guided enhancement. For example, on NegBench COCO, InternVL2.5 Acc score increases from 0.912 to 0.965. 
All the top Acc scores establish new SOTA performance on the benchmarks, achieved on recent frontier MLLMs without fine-tuning. Fusion makes changes of the predictions from both ILC and ELC. The comparison of the changes would also verify the trustworthiness of ILC and ELC even without gt annotation, see Suppl. Sec 4.4.


\noindent {\bf Confusion matrices between two channels}:
With gt annotation, we categorize the matching of two channels as 4 cases: WW (win-win), WL (ILC win but ELC lost), LW (ILC lost but ELC win), and LL (lost-lost), and generate the confusion matrix (see Suppl. Sec 2.3). It is observed that, if CR score is high, the percentage of WW is high. The relative difference between ILC and ELC leads to the difference of WL and LW. If ILC is stronger than ELC, the percentage of WL is larger than LW, and vice versa. 
Here, WL corresponds to FN (Inconsistent but ILC is Correct). CR is strongly correlated with WW and less affected by WL errors.

\noindent {\bf Ablation study on Lose Cascading Effects in ELC}: 100 samples are randomly selected from HC-RefCOCOg val set, and then GT nouns and boxes are added manually. ELC performance is evaluated on different VFMs, on the nouns extracted by Mistrial (LLM), on the Ground-Truth nouns annotated manually, and on the Ground-Truth boxes of related objects labeled manually. Full results are presented in Suppl Sec 4.3. 
In summary, the errors of noun extraction by LLM may cause 10\% gap and box allocation by cause another 10\% gap on HC-RefCOCOg. We also performed ablation study on adversarial images. We manually mask one critical object in each image on 100 images, and test with different models. As example, on Qwen3.0-VL, Acc50 of ILC drops from 91\% to 60\%, while Acc50 of ELC drops from 82\% to 41\%, and CR drops from 0.80 to 0.57. It is observed that ELC is more sensitive to absence of referred objects.

\noindent {\bf Adaption to new tasks}: For a new task, user knows critical facts and logic rules for safety of decisions. One may employ AI tools to select LLM, VFM, design prompt, and implement logic inference~\cite{10.1093/oso/9780198537465.001.0001,Huth_Ryan_2004} to build ELC for the task.

\section{Conclusion}\label{sec:conclusion}
Facing the uncertainty on reliability when deploying frontier MLLMs to new tasks, we proposed an Explicit Logic Channel for model validation and enhancement without any re-training or fine-tuning. Leveraging foundation models and logical reasoning~\cite{10.1093/oso/9780198537465.001.0001}, ELC produces decisions grounded on explicit facts and relations, providing a principled way for model validation. We investigated the effectiveness of proposed framework on three challenging benchmarks using 11 frontier MLLMs. 
The CR serves as reliable metric for model validation, selection and enhancement, with enhanced explainability and trustworthiness. The diversity of the three VLC tasks demonstrates the generality and flexibility of our approach in real-world applications. Extending ELC to more complex multimodal CoT reasoning tasks is a promising direction for future work.


\section*{Acknowledgements}
This research/project is supported by the National Research Foundation, Singapore under its National Large Language Models Funding Initiative (AISG Award No: AISG-NMLP-2024-004). Any opinions, findings and conclusions or recommendations expressed in this material are those
of the author(s) and do not reflect the views of National Research Foundation, Singapore. The authors would like to thank the anonymous reviewers for their constructive comments and valuable suggestions, which helped improve the quality and clarity of this paper.
%
%

\appendix

\begin{center}
    {\LARGE\bfseries Supplementary Material}
\end{center}

\vspace{1em}
\title{Supplementay Material \\
Explicit Logic Channel for Validation and Enhancement of MLLMs on Zero-Shot Tasks} 

\titlerunning{ELC for MLLM Validation and Enhancement}


\authorrunning{M.C. Leong, Y. Gu, H.L. Tan, L. Li, N.F. Chen}



In Section~\ref{sec:prompts}, we present all the prompts used for the VLC tasks and applied to the challenging benchmarks for evaluation. In Section~\ref{sec:cr-correction}, we present the statistics on raw experimental outcomes to verify the assumption of aligned fusion. In Section~\ref{sec:details}, we present the full experimental details beyond the main paper. In Section~\ref{sec:ablation}, we present ablation studies. Finally, in Section~\ref{sec:visual-evaluation}, we present visual examinations on inconsistent samples from each benchmark and the new findings.

\section{Prompt Descriptions}\label{sec:prompts}
To exploit the generalization of foundation models, such as LLM and MLLM, we employ prompts to extract semantic facts and relations for explicit logic reasoning. The details of the prompts are described in the following.

\subsection{Prompts for MLLM in ILC}
In ILC, the MLLM is prompted to make the prediction with the prompt templates provided by corresponding benchmarks. The full prompts can be found from their links, as presented in Subsection~\ref{sec:dataset-links}. 

\subsection{Prompts to LLM for language interpretation in ELC}
For MC-VQA task on NegBench, the prompt for LLM to extract objects present in or absent from the image is presented as the following: \\
\textit{“You are given a sentence. Return a dict of present nouns and absent nouns. Example:  `No book' means {`absent': [`book']}, `cake is not visible' means {`absent': [`cake']}. Note: Do not give any explanation. Output a complete JSON string with only 2 key: `present', `absent'.”}\\

For HC-REC task on HC-RefCOCOg, The prompt for extracting object nouns, as well as other potentially useful information is presented as follows: \\
\textit{
[INST] Extract the following information from the sentence:\\
1. A list of object nouns. No person noun, no scene noun.\\ 
Note:\\
1. Do not give any explaination.\\
2. For object nouns, include color of object, for example, ``orange shirt'', ``black shorts'', ``pink guitar''.\\
3. For object nouns, remove person nouns, such as: ``man'', ``group of students'', ``mother'', ``they'', ``tourists''.\\
Sentence: ``\{\}''\\
Output a complete JSON string with only single key: ``object\_nouns''\\
{[/INST]}}

For HC-REC task on HC-RefLoCo with long context query, the LLM is prompted to classify each sentence into three categories. The prompt is presented as the following:\\
\textit{
You are an expert linguistic annotator. Your task is to analyze sentences which contain long-context referring expressions for identifying one person in an image.\\
\#\#\# Task\\
Categorize the provided sentence into EXACTLY one of the following three categories:\\
1. **Essential Facts**: Information that directly describes the target person and associated objects. This includes their physical appearance (clothes, hair, age), their current actions, their name/identity (if a celebrity), or their specific posture.\\
   - *Example:* ``The man is wearing a dark pinstripe suit.''\\
   - *Example:* ``He appears to be engaged in the activity of washing someone's hair.''\\
2. **Non-Essential Facts**: Information that focuses on inanimate objects or details that do not describe the person themselves. Even if the person is interacting with the object, if the sentence focus is on the object's properties, it belongs here.\\
   - *Example:* ``The sizeable red fuel container is made of heavy plastic.''\\
   - *Example:* ``The framed certificate has a gold border.''\\
3. **Environment**: Information regarding the surroundings, the location, the environment, or other people in the scene used as a reference point.\\
   - *Example:* ``She is standing in line with other firefighters.''\\
   - *Example:* ``The scene appears to be a busy professional kitchen.''\\
\#\#\# Instruction\\
- Analyze the sentence focus.\\
- If a sentence contains a mix, prioritize ``Essential Facts'' if it helps identify the person, otherwise prioritize ``Environment''.\\
- Output ONLY the category name in JSON format.\\
\#\#\# Input Sentence:\\
``[SENTENCE]''\\
\#\#\# Output Format:
\{
  ``category'': ``'',
\}
}

\subsection{Prompt to VLM in ELC for HC-REC tasks}
For HC-REC tasks, if a CLIP-based model is used as VLM in ELC, each cropped person patch and referring text are fed to the VLM to obtain a score of visual-language matching. If an MLLM is used as the VLM, a prompt is presented as\\ 
\textit{question = ``Can you give a number between 0 and 1 representing similarity of the description `+ label.rstrip(`.') + ' and this image? The output is just value number using the same format.''}           

\section{Analysis of ILC and ELC relations}

\subsection{Logic Consistency and True Correction}\label{sec:cr-correction}

\begin{figure}[tb]
  \centering
  \includegraphics[width=0.9\columnwidth]{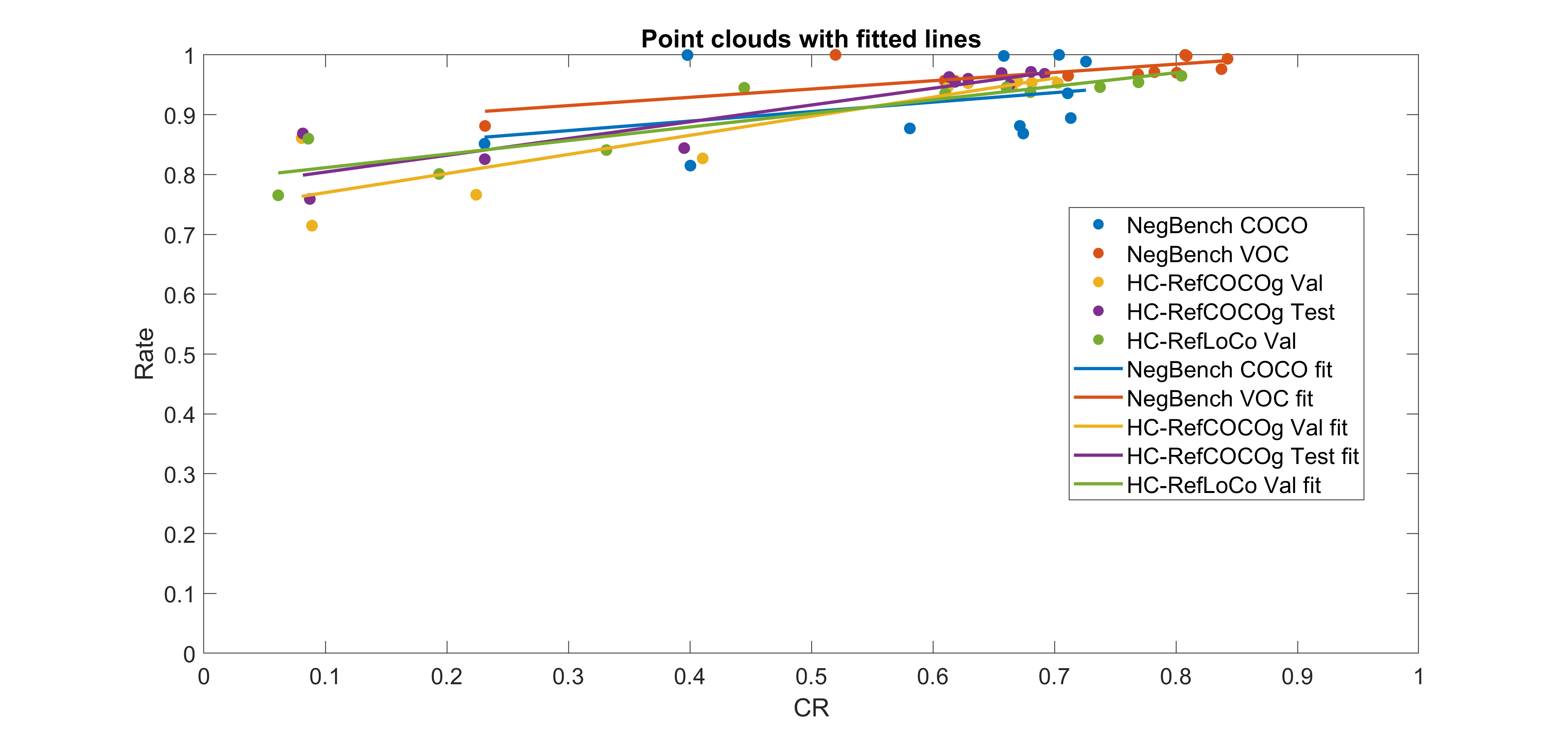}
  \caption{Correlation between CR scores and $CR_{gt}/CR$ ratios.}
  \label{fig:cr-correction}
\end{figure}
In principle, when the ILC and ELC predictions are logically consistent, the prediction is highly likely to be correct. To verify this assumption, we compute $CR_{gt}$ on samples that are both consistent and correct when given ground-truth. The ratio $CR_{gt}/CR$ is computed for every model across all tasks. The relations of CR scores and the ratios are plotted in Figure~\ref{fig:cr-correction}, where the cloud of the same color points are the results from the experiments on one benchmark, and the line of the same color is the linear fitting of the point cloud. It can be observed that the points are crowded over 80\%, especially when $CR$>0.5. From the fitted lines, one can observe that the larger the CR, the higher the ratio, indicating that higher consistency rate (CR) leads to higher correctness and reliability. 

\subsection{Logic Consistent but  Incorrect (not matching gt)}\label{sec:ci-analysis}

\begin{table}[th]
  \caption{Evaluation of CI (Consistent but Incorrect (not matching gt annotation)) cases on HC-RefCOCOg val, where the rates are obtained at IoU=0.5.}
  \label{tab:CI-on-HC-RefCOCOg_val}
  \centering
  \resizebox{0.7\textwidth}{!}{
  \begin{tabular}{l|c|c|c|c|l}
    \toprule
      & & \multicolumn{3}{c|}{Rates of CI categories} & \\ 
    \midrule     
    MLLM & CI Rate & Model Bias & ELC Error & GT Error & CR \\
    \midrule
    InternVL2.5-8B  & 0.0437 & 44.2\% & 9.6\% & 46.2\% & 0.6695 \\
    Qwen3.0-VL-8B  & 0.0473 & 32.2\% & 15.3\% & 52.2\% & 0.7027 \\
  \bottomrule
  \end{tabular}
  }
\end{table}
On the other hand, the cases of consistent but incorrect (not matching gt label) might raise concerns on CR, especially, they may cause false positives of trustworthy when gt labels not available. To investigate the risks of Consistent but Incorrect (CI) cases, we manually evaluate such cases on HC-RefCOCOg val set with InternVL-2.5 and Qwen3.0-VL, two MLLMs with high CR scores at IoU=0.5. On HC-RefCOCOg val set of 1776 test queries, Qwen3.0-VL generates 59 CI cases, and InternVL-2.5 generates 52 CI cases. Through manual examinations, it is found that CI cases can be classified into three categories, i.e., biased predictions due to shared training data, errors of ELC, and poor gt annotations. The error distributions are presented in Tables~\ref{tab:CI-on-HC-RefCOCOg_val}. It can be observed that, when Qwen3.0-VL is employed in ILC, the CR score is 0.703 (top 1), the rate of CI is 4.7\%, and of them, 52\% are caused by poor gt annotations, 32\% are related to bias of models, and only 15\% are caused by ELC errors, and when InternVL-2.5 is employed in ILC, the CR score is 0.670, the rate of CI is 4.4\%, and of them, 46\% are caused by poor gt annotations, 44\% are related to bias of models, and only 10\% are caused by ELC errors. The examples of the three categories of CI cases are presented in Figure~\ref{fig:ci-examples}. In summary, Once CR score is high, the CI risk is low, and mainly caused by gt noise and common bias of models.

\begin{figure}[tb]
  \centering
  \includegraphics[width=1.0\columnwidth]{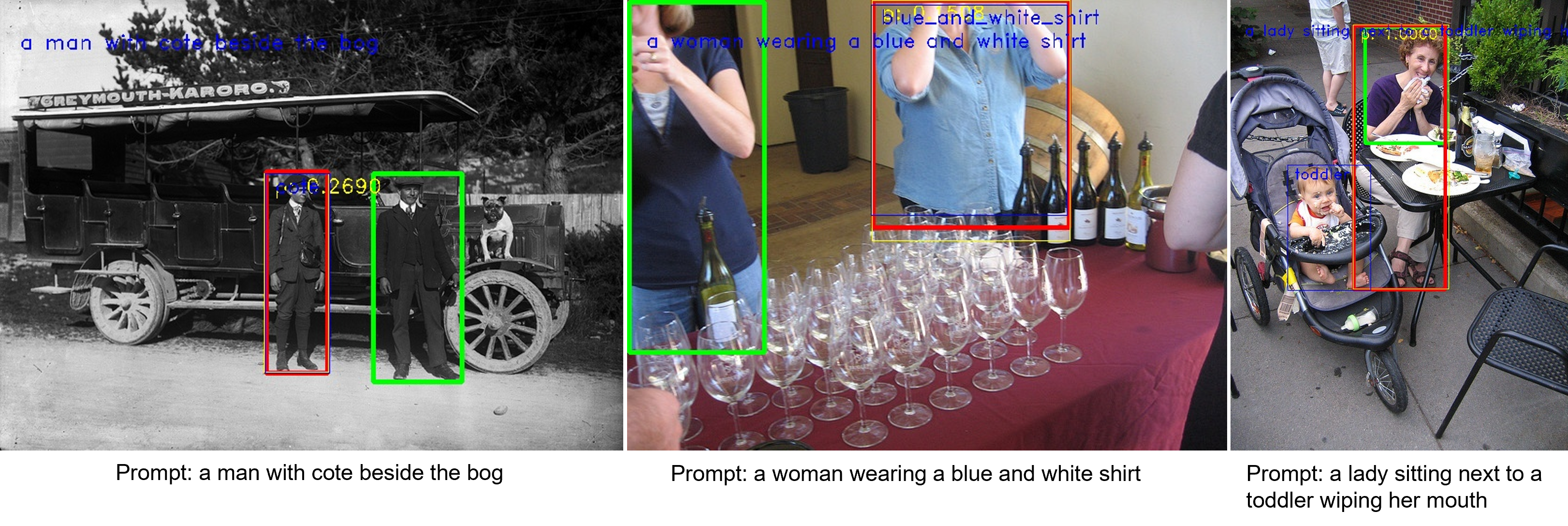}
  \caption{The examples of consistent-but-incorrect (CI) cases. In the image, the green box is the gt annotation, the red box is the prediction by MLLM in ILC, the yellow box is the prediction by ELC, and the blue boxes indicate the detected related objects on nouns. The left image is an example of CI case due to Biased Models, where both ILC and ELC predict the left person, but the gt person is the right one. The middle image shows an example of ELC Error, where the GroundingDINO might not be able to discriminant ``blue and white shirt'' with ``white blue shirt''. The right image shows an example of GT Error, where the gt box just includes the lady's upper body, but both ILC and ELC predict the full body box.}
  \label{fig:ci-examples}
\end{figure}

\subsection{Full confusion matrix}\label{sec:confusion-matrix}
When gt annotation is available, the joint outcomes of the two channels can be classified into four categories, {\em i.e.}, Win-Win (WW): ILC and ELC make the same prediction and that matches gt annotation, Win-Lost (WL): ILC predicts the correct answer but ELC fails, Lost-Win (LW): ELC predicts the correct answer but ILC fails, and Lost-Lost (LL): both ILC and ELC fail to predict the correct answer. The confusion matrix of the joint outcomes provides additional information on the performance of ILC and ELC, as well as their relations. On benchmarks of NegBench-COCO and HC-RefCOCOg val, we generate the confusion matrices on four representative MLLMs from the corresponding frontier families. The results are presented in Table~\ref{tab:confusion-matrices}.
\begin{table}[th]
  \caption{Confusion matrices on NegBench-COCO (upper block) and HC-RefCOCOg val at IoU=0.5 (lower block) with four representative MLLM from the four frontier families employed in ILC.}
  \label{tab:confusion-matrices}
  \centering
  \resizebox{0.6\textwidth}{!}{
  \begin{tabular}{l|c|c|cccc}
    \toprule
      & \multicolumn{2}{c|}{Performance} & \multicolumn{4}{c}{Confusion Matrix} \\ 
    \midrule     
    MLLM & Acc & CR & WW & WL & LW & LL \\
    \midrule
    LLaVA-1.5-13B  & 0.6206 & 0.5810 & 51.0\% & 11.2\% & 24.8\% &13.1\% \\
    Gemma-3-12B  & 0.7198 & 0.6715 & 59.2\% & 12.8\% & 16.6\% &11.4\% \\
    Qwen2.5-VL-7B  & 0.8165 & 0.7109 & 66.5\% & 15.1\% & 9.3\% & 9.1\% \\
    InternVL3.0-8B  & 0.9319 & 0.7259 & 71.8\% & 21.4\% & 4.0\% & 2.8\% \\
    ELC  & 0.7577 &  &  &  &  &  \\
  \bottomrule
    Gemma-3-12B  & 0.0884 & 0.0805 & 7.0\% & 1.2\% & 64.0\% &17.5\% \\
    LLaVA-1.5-13B  & 0.2190 & 0.2241 & 17.2\% & 3.8\% & 53.3\% &15.4\% \\
    InternVL3.5-8B  & 0.7798 & 0.6813 & 65.0\% & 12.4\% & 6.4\% & 5.9\% \\
    Qwen3.0-VL-8B  & 0.8035 & 0.7072 & 67.0\% & 12.6\% & 5.0\% & 5.3\% \\
    ELC  & 0.8074 &  &  &  &  &  \\
  \bottomrule
  \end{tabular}
  }
\end{table}

It is observed that, if CR score is high, the percentage of WW is high. The relative difference of accuracies between ILC and ELC leads to the difference of WL and LW. If ILC is stronger than ELC, the percentage of WL is larger than LW, and vice versa. As example, on NegBench-COCO, when ILC employs Qwen3.0-VL, WW is 72\%, WL is 21\%, LW is only 4\%, and LL is just 3\%, as Acc of ILC is 93\%, larger than 76\% of ELC. When LLaVA-1.5 is used in ILC, WW is 51\%, WL is 11\%, LW is 25\%, and LL is 13\%, as Acc of ILC is 62\%, less than 76\% of ELC. CR is strongly correlated with WW. Similar observations can be obtained on the confusion matrices on HC-RefCOCOg val with IoU=0.5.

\section{Experimental Details and Full Results}\label{sec:details}

\subsection{Links of Data Sources}\label{sec:dataset-links}
The three benchmark datasets can be prepared or downloaded following the links below: 
\begin{itemize}
\item NegBench benchmark: \url{https://github.com/m1k2zoo/negbench}
\item RefCOCOg benchmark: \url{https://github.com/lichengunc/refer}
\item HC-RefLoCo benchmark:\\ \url{https://github.com/ZhaoJingjing713/HC-RefLoCo}
\end{itemize}

\subsection{Implementation Details}
In our experiments, all the MLLM, LLM, VFM and VLM models are implemented in a server with two Nvidia RTX2080Ti GPUs.

The full results on NegBench have been presented in Table 1 in the main paper. Due to limited space in the main paper, only representative results on HC-RefCOCOg and HC-RefLoCo are presented. Here, we present the full results. The full results on HC-RefCOCOg are presented in Table~\ref{tab:HC-RefCOCOg_val} and Table~\ref{tab:HC-RefCOCOg_test}, including CR/50/75/90 obtained at IoU thresholds of 0.5, 0.75, and 0.9, and the enhanced performance on val set. In these tables, `RA' means Rank on $Acc$, and `RC' means Rank on the score of $CR$. CR50 can be used as the representation of CR/50/75/90. As reference, the accuracies of ELC are listed at the bottom rows of the tables. Although the accuracy of ELC is not the highest, but when combined with ILC, it could achieve better performance even surpassing the top ILC performance. As an example, ILC with Qwen3.0-VL-8B has achieved 0.8035/0.7494/0.5974 of $Acc$/50/75/90, when combined with ELC, the final performance reaches 0.8463/0.7849/0.6486 of $Acc_E$/50/75/90.

\begin{table}[th]
  \caption{HC-REF on HC-RefCOCOg val.}
  \label{tab:HC-RefCOCOg_val}
  \centering
  \resizebox{0.9\textwidth}{!}{
  \begin{tabular}{l|l|l|c|c|l}
    \toprule
    MLLM & $Acc$/50/75/90 & $CR$/50/75/90 & RA & RC & $Acc_E$/50/75/90 \\
    \midrule
    Gemma-3-12B  & .0884/.0039/.0011 & .0805/.0039/.0000 & 10 & 11 & .7466/.6869/.5715 \\
    InternVL2.0-8B  & .7506/.6571/.3857 & .6650/.5895/.3896 & 4 & 4 & .8356/.7720/.6177 \\
    InternVL2.5-8B  & .7646/.7100/.5168 & .6695/.6171/.4848 & 3 & 3 & .8423/.7809/.6391 \\
    InternVL3.0-8B  & .6993/.6199/.4032 & .6182/.5524/.3722 & 6 & 6 & .8215/.7613/.6143 \\
    InternVL3.5-8B  & .7798/.7354/.5968 & .6813/.6334/.5670 & 2 & 2 & .8446/.7838/.6515 \\
    LLaVA-1.5-13B  & .2190/.0693/.0118 & .2241/.0788/.0146 & 9 & 9 & .7551/.6841/.5636 \\
    LLaVA-1.6-13B  & .0760/.0068/.0011 & .0890/.0073/.0006 & 11 & 10 & .7432/.6858/.5698 \\
    QwenVL-7B-Chat  & .3908/.3609/.2996 & .4105/.3818/.3271 & 8 & 8 & .7810/.7247/.5997 \\
    Qwen2.0-VL-8B  & .7150/.6306/.4887 & .6289/.5529/.4724 & 5 & 5 & .8204/.7528/.6143 \\
    Qwen2.5-VL-8B  & .6841/.5907/.3947 & .6121/.5304/.3823 & 7 & 7 & .8311/.7618/.6188 \\
    Qwen3.0-VL-8B  & .8035/.7494/.5974 & .7027/.6503/.5794 & 1 & 1 & .8463/.7849/.6486 \\
    \midrule
    ELC  & .8074/.7412/.6278 &  & &  &  \\
  \bottomrule
  \end{tabular}
  }
\end{table}

\begin{table}[th]
  \caption{HC-REF on HC-RefCOCOg test.}
  \label{tab:HC-RefCOCOg_test}
  \centering
  \resizebox{0.9\textwidth}{!}{
  \begin{tabular}{l|l|l|c|c|l}
    \toprule
    MLLM & $Acc$/50/75/90 & $CR$/50/75/90 & RA & RC & $Acc_E$/50/75/90 \\
    \midrule
    Gemma-3-12B  & .0884/.0043/.0009 & .0815/.0040/.0009 & 10 & 11 & .7515/.6899/.5871 \\
    InternVL2.0-8B  & .7553/.6715/.4077 & .6631/.5885/.4008 & 4 & 3 & .8413/.7766/.6372 \\
    InternVL2.5-8B  & .7748/.7216/.5333 & .6565/.5992/.4843 & 3 & 4 & .8503/.7912/.6594 \\
    InternVL3.0-8B  & .7132/.6386/.4273 & .6133/.5439/.3746 & 6 & 7 & .8367/.7705/.6329 \\
    InternVL3.5-8B  & .8079/.7610/.6098 & .6807/.6355/.5586 & 2 & 2 & .8537/.7956/.6700 \\
    LLaVA-1.5-13B  & .2358/.0757/.0153 & .2312/.0717/.0132 & 9 & 9 & .7682/.6945/.5859 \\
    LLaVA-1.6-13B  & .0824/.0052/.0023 & .0872/.0086/.0046 & 11 & 10 & .7518/.6931/.5911 \\
    QwenVL-7B-Chat  & .3907/.3645/.3046 & .3953/.3565/.2951 & 8 & 8 & .7803/.7242/.6162 \\
    Qwen2.0-VL-8B  & .7348/.6473/.5073 & .6289/.5572/.4780 & 5 & 5 & .8330/.7685/.6430 \\
    Qwen2.5-VL-8B  & .7089/.6196/.3904 & .6182/.5379/.3942 & 7 & 6 & .8385/.7757/.6421 \\
    Qwen3.0-VL-8B  & .8183/.7765/.6323 & .6919/.6467/.5788 & 1 & 1 & .8560/.7979/.6726 \\
    \midrule
    ELC  & .8005/.7434/.6337 &  & &  &  \\
  \bottomrule
  \end{tabular}
  }
\end{table}

The full results on HC-RefLoCo are presented in Table~\ref{tab:HC-RefLoCo_val} and Table~\ref{tab:HC-RefLoCo_test}, including $Acc$/50/75/90 and CR/50/75/90 obtained at IoU thresholds of 0.5, 0.75, and 0.9, as well as the enhanced performance on val set. $Acc$50 and CR50 can be used as representations of $Acc$/50/75/90 and CR/50/75/90 for validation of ranking. Meanwhile, the significant improvements of $Acc$90 after enhancement indicate that the ELC is useful in improving the accuracy of localization for HC-REC tasks. The accuracies of ELC are listed at the bottom rows of the tables as reference.

\begin{table}[th]
  \caption{HC-REF on HC-RefLoCo val.}
  \label{tab:HC-RefLoCo_val}
  \centering
  \resizebox{0.9\textwidth}{!}{
  \begin{tabular}{l|l|l|l|c|c|l|l}
    \toprule
    MLLM & $Acc$/50/75/90 & $mAcc$ & $CR$/50/75/90 & RA & RC & $Acc_E$/50/75/90 & $mAcc_E$ \\
    \midrule
    Gemma-3-12B  & .0921/.0040/.0001 & .0219 & .0860/.0031/.0001 & 10 & 10 & .4266/.3525/.2952 & .3496 \\
    InternVL2.0-8B  & .7592/.5769/.2889 & .5403 & .6802/.5251/.2907 & 4 & 4 & .8190/.6804/.4471 & .6453 \\
    InternVL2.5-8B  & .5039/.2921/.1504 & .3048 & .4447/.2628/.1454 & 7 & 7 & .6650/.5132/.3734 & .5077 \\
    InternVL3.0-8B  & .6817/.4972/.2638 & .4790 & .6100/.4542/.2622 & 6 & 6 & .7662/.6272/.4323 & .6054 \\
    InternVL3.5-8B  & .8832/.7105/.4232 & .6718 & .7691/.6359/.4172 & 2 & 2 & .8935/.7600/.5289 & .7249 \\
    LLaVA-1.5-13B  & .1906/.0609/.0100 & .0821 & .1936/.0608/.0118 & 9 & 9 & .4214/.3240/.2522 & .3270 \\
    LLaVA-1.6-13B  & .0626/.0024/.0001 & .0140 & .0612/.0030/.0005 & 11 & 11 & .3265/.2730/.2316 & .2703 \\
    QwenVL-7B-Chat  & .3284/.2899/.2028 & .2687 & .3313/.2948/.2334 & 8 & 8 & .4874/.4442/.3496 & .4217 \\
    Qwen2.0-VL-8B  & .8396/.6638/.4610 & .6532 & .7375/.6001/.4604 & 3 & 3 & .8582/.7260/.5496 & .7076 \\
    Qwen2.5-VL-8B  & .7502/.5657/.2719 & .5283 & .6604/.5132/.2783 & 5 & 5 & .8208/.6816/.4431 & .6460 \\
    Qwen3.0-VL-8B  & .9412/.8534/.6169 & .7992 & .8045/.7495/.6189 & 1 & 1 & .9386/.8580/.6523 & .8114 \\
    \midrule
    ELC  & .8044/.7543/.6258 & .7225 & & & &  \\    
  \bottomrule
  \end{tabular}
  }
\end{table}

\begin{table}[th]
  \caption{HC-REF on HC-RefLoCo test.}
  \label{tab:HC-RefLoCo_test}
  \centering
  \resizebox{0.9\textwidth}{!}{
  \begin{tabular}{l|l|l|l|c|c|l|l}
    \toprule
    MLLM & $Acc$/50/75/90 & $mAcc$ & $CR$/50/75/90 & RA & RC & $Acc_E$/50/75/90 & $mAcc_E$ \\
    \midrule
    Gemma-3-12B  & .0889/.0038/.0002 & .0207 & .0815/.0034/.0001 & 10 & 10 & .4315/.3637/.3065 & .3575 \\
    InternVL2.0-8B  & .7534/.5722/.2849 & .5350 & .6860/.5344/.4472 & 4 & 4 & .8120/.6750/.4472 & .6412 \\
    InternVL2.5-8B  & .4970/.2917/.1514 & .3041 & .4478/.2681/.1506 & 7 & 7 & .6618/.5228/.3827 & .5144 \\
    InternVL3.0-8B  & .6742/.4889/.2573 & .4720 & .6095/.4547/.2624 & 6 & 6 & .7630/.6298/.4379 & .6071 \\
    InternVL3.5-8B  & .8866/.7121/.4210 & .6733 & .7746/.6404/.4156 & 2 & 2 & .9003/.7679/.5359 & .7323 \\
    LLaVA-1.5-13B  & .1957/.0604/.0107 & .0834 & .1966/.0620/.0122 & 9 & 9 & .4304/.3338/.2601 & .3353 \\
    LLaVA-1.6-13B  & .0621/.0022/.0002 & .0144 & .0589/.0021/.0004 & 11 & 11 & .3335/.2853/.2391 & .2797 \\
    QwenVL-7B-Chat  & .3314/.2900/.2034 & .2699 & .3332/.2974/.2332 & 8 & 8 & .4959/.4538/.3570 & .4303 \\
    Qwen2.0-VL-8B  & .8451/.6686/.4549 & .6542 & .7449/.6001/.4621 & 3 & 3 & .8652/.7355/.5473 & .7123 \\
    Qwen2.5-VL-8B  & .7522/.5665/.2706 & .5288 & .6650/.5157/.2721 & 5 & 5 & .8260/.6905/.4509 & .6526 \\
    Qwen3.0-VL-8B  & .9418/.8557/.6280 & .8024 & .8128/.7565/.6261 & 1 & 1 & .9388/.8607/.6602 & .8143 \\
    \midrule
    ELC  & .8146/.7661/.6411 & .7342 & & & &  \\    
  \bottomrule
  \end{tabular}
  }
\end{table}

\section{Ablation Studies}\label{sec:ablation}

In the proposed ELC, foundation models of LLM and VLM are employed. We conduct ablation studies on the effectiveness and sensitivities of these models. The results are presented in the following subsections.

\subsection{Ablation study on LLM in ELC}

In ELC for both MC-VQA and HC-REC tasks, one representative LLM is employed as the language parser to extract basic concepts of visual facts and relations associated with the text query. In the following ablation studies, we implement ELC with two representative LLMs, \ie, Mistral~\cite{jiang2023mistral7b} and Qwen3.0~\cite{qwen3technicalreport}, and perform experiments on the three benchmarks. We compare the results with the two LLMs in ELC for evaluation. 

For the task of MC-VQA on NegBench, LLM is prompted to extract positive and negative objects from the text query. Since the COCO dataset is more challenging than VOC2007, as shown in Table 1 in the main paper, we perform ablation experiments on the COCO set. The comparison is presented in Table~\ref{tab:LLM-NegBench}, where the results with five MLLMs are presented, representing models with top, moderate and low level performance. The performance of ELC for corresponding LLMs are listed at the bottom line of the table. When comparing the corresponding scores of $Acc$, $CR$ and $Acc_E$ obtained with Mistral and Qwen3.0, respectively, one can observe that the differences are very small, less than 1\% level.
\begin{table}[th]
  \caption{Ablation experiments on LLM for MC-VQA task on NegBench COCO set.}
  \label{tab:LLM-NegBench}
  \centering
  \begin{tabular}{l|l|l|l|l|l|l}
    \toprule
     LLM (in ELC) & \multicolumn{3}{c|}{Mistral} & \multicolumn{3}{c}{Qwen3.0} \\ 
    \midrule     
    MLLM (in ILC) & $Acc$ & $CR$ & $Acc_E$ & $Acc$ & $CR$ & $Acc_E$ \\
    \midrule
    InternVL2.0-8B & 0.4878 & 0.3980 & 0.8434 & 0.4878 & 0.3999 & 0.8458 \\    
    InternVL3.0-8B & 0.9319 & 0.7259 & 0.9557 & 0.9319 & 0.7296 & 0.9557 \\
    LLaVA-1.5-13B & 0.6206 & 0.5810 & 0.6206 & 0.6206 & 0.5815 & 0.6206 \\
    QwenVL-7B-Chat & 0.2426 & 0.2310 & 0.6245 & 0.2426 & 0.2305 & 0.6272 \\
    Qwen3.0-VL-8B & 0.7753 & 0.7134 & 0.7758 & 0.7753 & 0.7158 & 0.7758  \\
    \midrule
    ELC & 0.7577 &  &  & 0.7614 &  &  \\
  \bottomrule
  \end{tabular}
\end{table}

For the task of HC-REC on HC-RefCOCOg, LLM is prompted to extract the associated objects and attributes to localize the referred person. We perform ablation experiments using Mistral and Qwen3.0 as the LLM in ELC. Again, five MLLMs are selected to represent models with top, moderate and low level performance. The comparison on $CR$ and enhanced performance is presented in Table~\ref{tab:LLM_HC-RefCOCOg}. The performance of ELC for corresponding LLMs are listed at the bottom of the table. When comparing the corresponding scores of $CR$ and $Acc_E$ obtained with Mistral and Qwen3.0, respectively, one can observe that the differences are very small, less than 1\% level.

\begin{table}[th]
  \caption{Ablation experiments on LLM for HC-REC task on HC-RefCOCOg val set.}
  \label{tab:LLM_HC-RefCOCOg}
  \centering
  \resizebox{1.0\textwidth}{!}{
  \begin{tabular}{l|l|l|l|l}
    \toprule
     LLM (in ELC) & \multicolumn{2}{c|}{Mistral} & \multicolumn{2}{c}{Qwen3.0} \\ 
    \midrule     
    MLLM (in ILC) & $CR$/50/75/90 & $Acc_E$/50/75/90 &  $CR$/50/75/90 & $Acc_E$/50/75/90 \\
    \midrule
    InternVL2.0-8B & 0.6650/0.5895/0.3896 & 0.8356/0.7720/0.6177 & 0.6706/0.5923/0.3896 & 0.8333/0.7691/0.6137 \\    
    InternVL3.0-8B & 0.6182/0.5524/0.3722 & 0.8215/0.7613/0.6143 & 0.6244/0.5586/0.3744 & 0.8198/0.7590/0.6109 \\
    LLaVA-1.5-13B & 0.2241/0.0788/0.0146 & 0.7551/0.6841/0.5636 & 0.2258/0.0794/0.0152 & 0.7528/0.6813/0.5597 \\
    QwenVL-7B-Chat & 0.4105/0.3818/0.3271 & 0.7810/0.7247/0.5997 & 0.4116/0.3840/0.3283 & 0.7787/0.7218/0.5957 \\
    Qwen3.0-VL-8B & 0.7027/0.6503/0.5794 & 0.8463/0.7849/0.6486 & 0.7123/0.6561/0.5856 & 0.8440/0.7821/0.6447  \\
    \midrule
    ELC ($Acc$) &  & 0.8074/0.7412/0.6278 &  & 0.7996/0.7393/0.6199 \\
  \bottomrule
  \end{tabular}
  }
\end{table}

\begin{table}[th]
  \caption{Ablation experiments on LLM for HC-REC task on HC-RefLoCo val set.}
  \label{tab:LLM_HC-RefLoCo}
  \centering
  \resizebox{0.9\textwidth}{!}{
  \begin{tabular}{l|l|l|l|l|l|l}
    \toprule
    MLLM(ILC) & LLM & $Acc$/50/75/90 & $mAcc$ & $CR$/50/75/90 & $Acc_E$/50/75/90 & $mAcc_E$ \\
    \midrule
    LLaVA-1.5-13B & Qwen3.0 & .1906/.0609/.0100 & .0821 & .1936/.0608/.0118 & .4214/.3240/.2522 & .3270 \\
    LLaVA-1.5-13B & Mistral & .1906/.0609/.0100 & .0821 & .1935/.0609/.0120 & .4123/.3153/.2403 & .3169 \\
    \midrule
    InternVL3.0-8B & Qwen3.0 & .6817/.4972/.2638 & .4790 & .6100/.4542/.2622 & .7662/.6272/.4323 & .6054 \\
    InternVL3.0-8B & Mistral & .6817/.4972/.2638 & .4790 & .6132/.4566/.2629 & .7680/.6327/.4361 & .6085 \\
    \midrule    
    Qwen3.0-VL-8B & Qwen3.0 & .9412/.8534/.6169 & .7992 & .8045/.7495/.6189 & .9386/.8580/.6523 & .8114 \\
    Qwen3.0-VL-8B & Mistral & .9412/.8534/.6169 & .7992 & .8094/.7536/.6231 & .9383/.8564/.6501 & .8099 \\
    \midrule    
    ELC & Qwen3.0 & .8044/.7543/.6258 & .7225 &  &  & \\
    ELC & Mistral & .8110/.7603/.6317 & .7285 &  &  & \\
  \bottomrule
  \end{tabular}
  }
\end{table}

For the task of HC-REC on long context query, the LLM is employed as a language parser to classify each sentence into three categories related to localization of the referred person. In the results presented in main paper, Qwen3.0 is used as the LLM in ELC. To study the sensitivity of ELC on LLM, we also tested with Mistral as the LLM in ELC on HC-RefLoCo. From the results in Table 3 in the main paper, we select three representative MLLMs in ILC, with top, moderate, and low performance with Qwen3.0. The comparison of the performance between Qwen3.0 and Mistral is presented in Table~\ref{tab:LLM_HC-RefLoCo}. From the results, it is observed that the differences between the two models are very small. 

The observations from the three ablation experiments on LLM in ELC demonstrate that our proposed ELC framework is less sensitive to the selection of LLM as language parser.

\subsection{Ablation study on VLM in ELC}
For tasks of HC-REC, in ELC, a VLM is employed to provide visual-language matching probability. To evaluate the effectiveness of VLM in ELC, we investigate two representative VLMs, \ie, EvaCLIP on basic CLIP model and InternVL2.0 on LLM backbone. The comparison of ELC with EvaCLIP and InternVL2.0 as VLM on HC-RefCOCOg and HC-RefLoCo are presented in Table~\ref{tab:VLM_HC-RefCOCOg} and Table~\ref{tab:VLM_HC-RefLoCo}, respectively, where the column `R' means Rank based on $CR$50 on the left column. The ELC performance with EvaCLIP and InternVL2.0 as VLM are listed at the bottom of each table. The ELC performance is moderate as compared to the 11 MLLMs used in ILC. There is also a large gap between the ELC performance on EvaCLIP and InternVL2.0, \ie, larger that 10\% difference. However, such difference has almost no effect on the ranking of $CR$ obtained with ELC. 
When integrating ELC with ILC in the aligned fusion, the performance can be enhanced to be close to the MLLMs in ILC (\ie, <2-3\% mostly). These observations indicate that the effectiveness of model validation and enhancement is less affected by the selection of VLM in ELC.

\begin{table}[th]
  \caption{Ablation experiments on VLM for HC-REC task on HC-RefCOCOg val set.}
  \label{tab:VLM_HC-RefCOCOg}
  \centering
  \resizebox{1.0\textwidth}{!}{
  \begin{tabular}{l|l|l|l|l|l|l}
    \toprule
     VLM & \multicolumn{3}{c|}{EvaCLIP} & \multicolumn{3}{c}{InternVL2.0} \\ 
    \midrule     
    MLLM & $CR$/50/75/90 & R & $Acc_E$/50/75/90 &  $CR$/50/75/90 & R & $Acc_E$/50/75/90 \\
    \midrule
    Gemma-3-12B & .0805/.0039/.0000 & 11 & .7466/.6869/.5715 & .0760/.0039/.0006 & 11 & .5417/.4583/.3784 \\
    InternVL2.0-8B & .6650/.5895/.3896 & 4 & .8356/.7720/.6177 & .6115/.5394/.3589 & 4 & .8384/.7579/.5743 \\    
    InternVL2.5-8B & .6695/.6171/.4848 & 3 & .8423/.7809/.6391 & .6126/.5614/.4448 & 3 & .8514/.7872/.6194 \\
    InternVL3.0-8B & .6182/.5524/.3722 & 6 & .8215/.7613/.6143 & .5535/.4927/.3316 & 7 & .8136/.7387/.5698 \\
    InternVL3.5-8B & .6813/.6334/.5670 & 2 & .8446/.7838/.6515 & .6227/.5766/.5175 & 2 & .8542/.7922/.6532 \\
    LLaVA-1.5-13B & .2241/.0788/.0146 & 9 & .7551/.6841/.5636 & .2038/.0693/.0107 & 9 & .5901/.4797/.3761 \\
    LLaVA-1.6-13B & .0890/.0073/.0006 & 10 & .7432/.6858/.5698 & .0878/.0079/.0017 & 10 & .5175/.4538/.3744 \\
    QwenVL-7B-Chat & .4105/.3818/.3271 & 8 & .7810/.7247/.5997 & .3823/.3514/.2996 & 8 & .6548/.6036/.5084 \\
    Qwen2.0-VL-7B & .6289/.5529/.4724 & 5 & .8204/.7528/.6143 & .5777/.5146/.4386 & 5 & .8198/.7348/.5997 \\
    Qwen2.5-VL-7B & .6121/.5304/.3823 & 7 & .8311/.7618/.6188 & .5693/.5000/.3654 & 6 & .8046/.7275/.5687 \\
    Qwen3.0-VL-8B & .7027/.6503/.5794 & 1 & .8463/.7849/.6486 & .6408/.5952/.5338 & 1 & .8682/.7979/.6543  \\
    \midrule
    ELC ($Acc$) &  &  & .8074/.7412/.6278 &  &  & .6948/.6413/.5411 \\
  \bottomrule
  \end{tabular}
  }
\end{table}

\begin{table}[th]
  \caption{Ablation experiments on VLM for HC-REC task on HC-RefLoCo val set.}
  \label{tab:VLM_HC-RefLoCo}
  \centering
  \resizebox{1.0\textwidth}{!}{
  \begin{tabular}{l|l|l|l|l|l|l|l|l}
    \toprule
     VLM & \multicolumn{4}{c|}{InternVL2.0} & \multicolumn{4}{c}{EvaCLIP} \\ 
    \midrule     
    MLLM & $CR$/50/75/90 & R & $Acc_E$/50/75/90 & $mAcc_E$ & $CR$/50/75/90 & R & $Acc_E$/50/75/90 & $mAcc_E$ \\
    \midrule
    Gemma-3-12B & .0860/.0031/.0001 & 10 & .4266/.3525/.2952 & .3496 & .0726/.0033/.0002 & 10 & .3900/.3133/.2539 & .3099 \\
    InternVL2.0-8B & .6802/.5251/.2907 & 4 & .8190/.6804/.4471 & .6453 & .5782/.4476/.2492 & 4 & .8049/.6639/.4220 &.6269 \\    
    InternVL2.5-8B & .4447/.2628/.1454 & 7 & .6650/.5132/.3734 & .5077 & .3733/.2223/.1232 & 7 & .6437/.4916/.3436 & .4834 \\
    InternVL3.0-8B & .6100/.4542/.2622 & 6 & .7662/.6272/.4323 & .6054 & .5220/.3840/.2204 & 6 & .7487/.6109/.4070 & .5861 \\
    InternVL3.5-8B & .7691/.6359/.4172 & 2 & .8935/.7600/.5289 & .7249 & .6461/.5336/.3503 & 2 & .8800/.7444/.5104 & .7078 \\
    LLaVA-1.5-13B & .1936/.0608/.0118 & 9 & .4214/.3240/.2522 & .3270 & .1850/.0585/.0113 & 9 & .4095/.3106/.2321 & .3115 \\
    LLaVA-1.6-13B & .0612/.0030/.0005 & 11 & .3265/.2730/.2316 & .2703 & .0592/.0026/.0005 & 11 & .2868/.2370/.1899 & .2312 \\
    QwenVL-7B-Chat & .3313/.2948/.2334 & 8 & .4874/.4442/.3496 & .4217 & .3152/.2783/.2192 & 8 & .4775/.4371/.3362 & .4111 \\
    Qwen2.0-VL-7B & .7375/.6001/.4604 & 3 & .8582/.7260/.5496 & .7076 & .6335/.5121/.3956 & 3 & .8438/.7085/.5299 & .6901 \\
    Qwen2.5-VL-7B & .6604/.5132/.2783 & 5 & .8208/.6816/.4431 & .6460 & .5548/.4265/.2317 & 5 & .8107/.6675/.4161 & .6286 \\
    Qwen3.0-VL-8B & .8045/.7495/.6189 & 1 & .9386/.8580/.6523 & .8114 & .6789/.6308/.5238 & 1 & .9218/.8401/.6336 & .7926  \\
    \midrule
    ELC ($Acc$) &  &  & .8044/.7543/.6258 & .7225 &  &  & .6811/.6363/.5269 & .6090 \\
  \bottomrule
  \end{tabular}
  }
\end{table}

\subsection{Ablation study on Lose Cascading Effects in ELC}

To understand the effects of loses with LLM and VFM in ELC, ablation study is performed. 100 samples are randomly selected from HC-RefCOCOg val set, and then GT nouns and boxes are added manually. The evaluations are perform with variation of VFMs ({\em i.e.}, GroundingDINO and SAM3~\cite{}), nouns from LLM (Mistral) and gt annotation, and object boxes predicted by VFM and gt annotation. In addition, to evaluate the effects of the loses on CR and enhanced accuracy, the results with Qwen-2.5-VL and Qwen-3.0-VL employed in ILC are also presented. The full results are presented in Table~\ref{tab:Error-Cascading}. 
\begin{table}[th]
  \caption{Ablation experiments on Error Cascading Effects in ELC on HC-RefCOCOg val set, as well as the effects on CR and enhanced accuracy with Qwen-2.5-VL and Qwen-3.0-VL, where the results are presented at IoU as .50/.75/.90.}
  \label{tab:Error-Cascading}
  \centering
  \resizebox{0.9\textwidth}{!}{
  \begin{tabular}{l|c|c|c|c|c}
    \toprule
      & \multicolumn{2}{c|}{LLM(noun) to VFM} & \multicolumn{2}{c|}{GT(noun) to VFM} & GT(box) to LR \\ 
    \midrule     
     & GroundingDINO & SAM3 & GroundingDINO & SAM3 & LR \\
    \midrule
    ELC & 0.82/0.74/0.59 & 0.74/0.68/0.54 & 0.93/0.93/0.93 & 0.94/0.94/0.94 & 0.98/0.98/0.98 \\
    \midrule
    CR (w Qwen2.5) & 0.71/0.61/0.37 & 0.65/0.51/0.31 & 0.72/0.57/0.28 & 0.70/0.55/0.26 & 0.72/0.57/0.27 \\    
    $Acc_E$ (w Qwen2.5) & 0.86/0.77/0.58 & 0.84/0.75/0.56 & 1.00/1.00/1.00 & 1.00/1.00/1.00 & 1.00/1.00/1.00 \\
    \midrule
    CR (w Qwen3.0) & 0.83/0.74/0.59 & 0.74/0.68/0.54 & 0.93/0.93/0.93 & 0.94/0.94/0.94 & 0.98/0.98/0.98 \\
    $Acc_E$ (w Qwen3.0) & 0.86/0.76/0.61 & 0.86/0.78/0.64 & 0.99/0.99/0.99 & 0.99/0.99/0.99 & 0.99/0.99/0.99 \\
  \bottomrule
  \end{tabular}
  }
\end{table}
To evaluate ELC performance on different VFMs on the nouns extracted by Mistrial (LLM), ELC is run on GDINO and SAM3 (VFMs). Acc50 of ELC on GDINO is 82\% and that on SAM3 is 74\%, matching the observations in HC-RefLoCo paper~\cite{NEURIPS2024_80f0cd03}. When GT nouns are applied to replace the outputs of Mistrial, Acc50 of ELC on GDINO is increased to 93\%, and that on SAM3 is increased to 94\%. Finally, when GT boxes are applied to replace the outputs of VFMs, Acc50 of ELC is increased to 98\%, where one failure is due to heavy overlapping of boxes and another requires reasoning over observations. Hence, the errors of noun extraction by LLM may cause 10\% gap and box allocation by cause another 10\% gap on HC-RefCOCOg. Similar effects on CR and enhanced accuracy with two Qwen MLLMs are also observed on the table.

\subsection{Ablation study on the benefit of fusion}
The results in main paper and full results in Table~\ref{tab:HC-RefCOCOg_val}, Table~\ref{tab:HC-RefCOCOg_test}, Table~\ref{tab:HC-RefLoCo_val} and Table~\ref{tab:HC-RefLoCo_test} show that Fusion improves the Acc beyond both ILC and ELC. In additional, Fusion makes the changes on the predictions from both ILC and ELC. The number of changes made in Fusion on each channel may reveal which channel performs better even without gt annotation. An ablation study on NegBench-VOC2007 with InternVL-2.0 and 2.5 is performed. The details are presented in Table~\ref{tab:Fusion-Benefit}. On this benchmark dataset, The ELC's Acc is 86.8\%. When InternVL2.0 is employed in ILC, the Acc of ILC is 58.7\%, which is much lower than the Acc from ELC. Fusion increases Acc to 93.5\%. We observe that Fusion changed 41.3\% of predictions by ILC and 6.7\% of predictions by ELC, indicating that ELC is more accurate and reliable than ILC on this dataset. When InternVL2.5 is employed in ILC, the Acc of ILC is 92.0\%, over that of ELC. Fusion changed 8.0\% of predictions by ILC and 11.3\% of predictions by ELC, indicating that ILC might perform better than ELC. This new finding reveals an additional benefit of Fusion for applications without gt annotation.
\begin{table}[th]
  \caption{Ablation study on the benefit of fusion on NegBench-VOC2007 with InternVL2.0 and 2.5, where $Acc_m$ means the accuracy of the MLLM on ILC, $Acc_{LR}$ means the accuracy of Logic Reasoning on ELC, $Acc_E$ means the enhanced accuracy by fusion, `Changes on ILC' means the number and rate of changes made in fusion on predictions from ILC, and `Changes on ELC' means the number and rate of changes made in fusion on the predictions from ELC .}
  \label{tab:Fusion-Benefit}
  \centering
  \resizebox{0.8\textwidth}{!}{
  \begin{tabular}{l|l|l|l|l|l|l}
    \toprule
    MLLM & $Acc_m$ & $Acc_{LR}$ & CR & $Acc_E$ & Changes on ILC & Changes on ELC \\
    \midrule
    InternVL2.0-8B & 0.5868 & 0.8684 & 0.5198 & 0.9352 & 2078/0.4130 & 338/0.0672 \\    
    InternVL2.5-8B & 0.9201 & 0.8684 & 0.8076 & 0.9809 & 402/0.0799 & 566/0.1125 \\
  \bottomrule
  \end{tabular}
  }
\end{table}



\section{Visual Examinations and New Findings}\label{sec:visual-evaluation}

One critical limitation of MLLMs is the uncertainty of their prediction when there is no ground-truth. With explicit logic consistency, we may effectively address such concern by manually examining the inconsistent samples even without gt annotation. The inconsistencies may be caused by:
\begin{itemize}
\item Hallucination: MLLM makes a prediction on nonexistent object or fails to see the referred object;
\item Ambiguity: MLLM makes a decision on neither sufficient nor necessary condition, or gt annotation is incorrect;
\item Weakness of FMs in ELC: LLM fails to extract the correct concept, or VFM fails to detect the relevant objects.
\end{itemize}
Examples of such errors observed on explicit visual evidence in inconsistent samples are presented in the following.

\begin{figure*}[h]
\centering
\includegraphics[width=1.0\columnwidth]{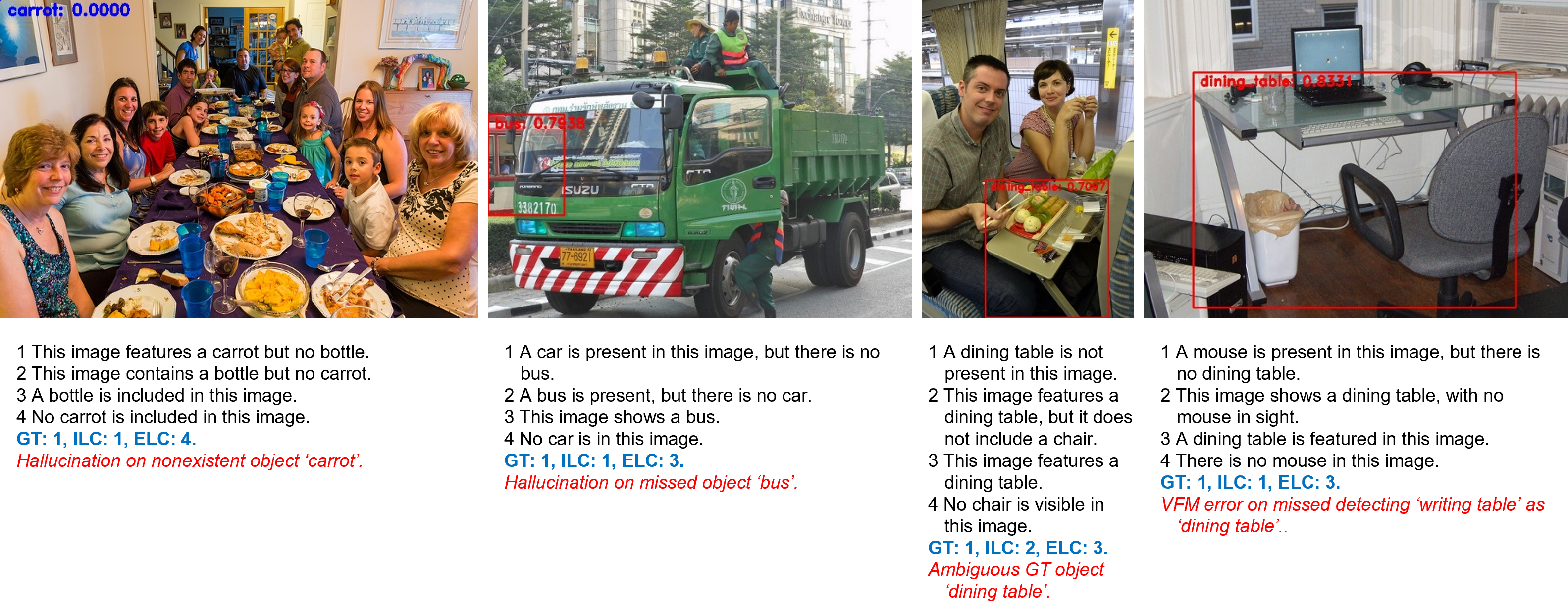}
\caption{Inconsistent examples from NegBench, where the last red text under each example indicates the cause of error on explicit visual evidence.}
\label{Fig:Examples_NegBench}
\end{figure*}

\begin{figure*}[h]
\centering
\includegraphics[width=1.0\columnwidth]{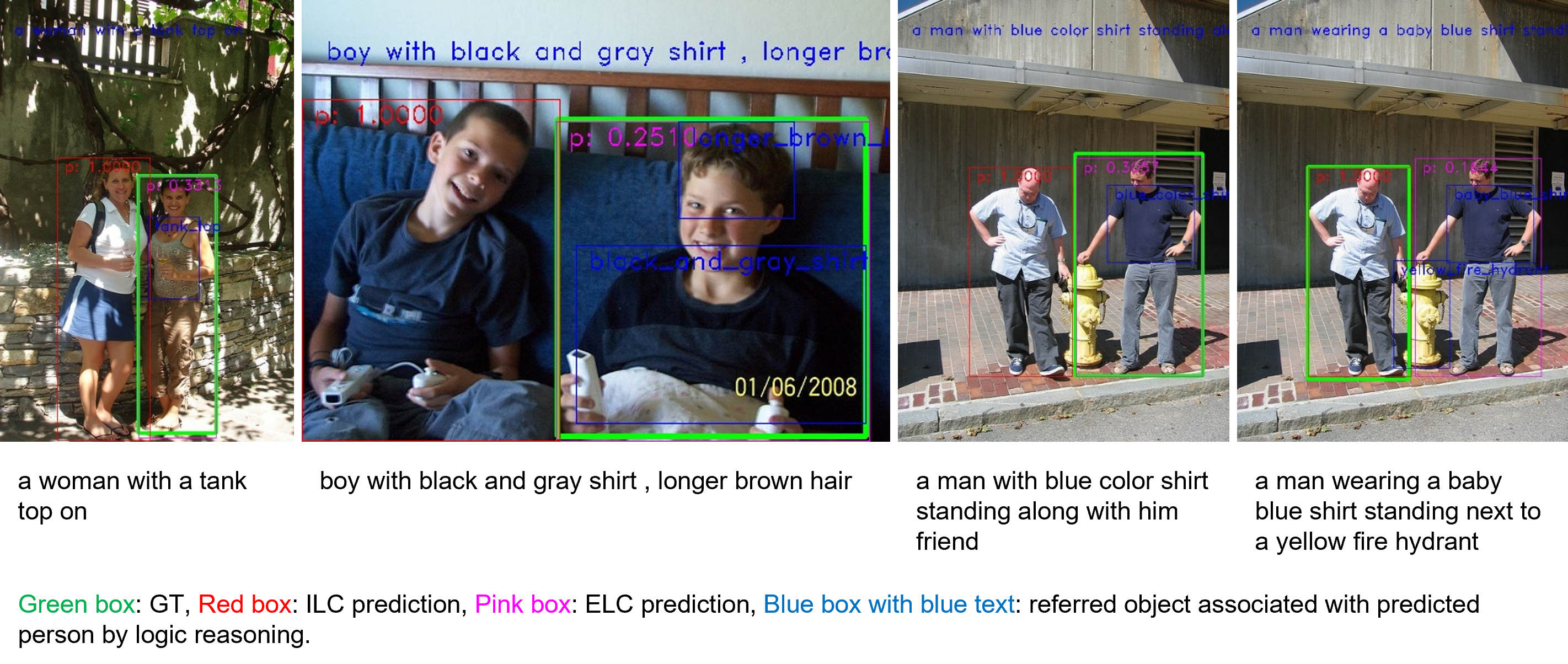}
\caption{Inconsistent examples from HC-RefCOCOg.}
\label{Fig:Examples_HC-RefCOCOg}
\end{figure*}

\begin{figure*}[h]
\centering
\includegraphics[width=1.0\columnwidth]{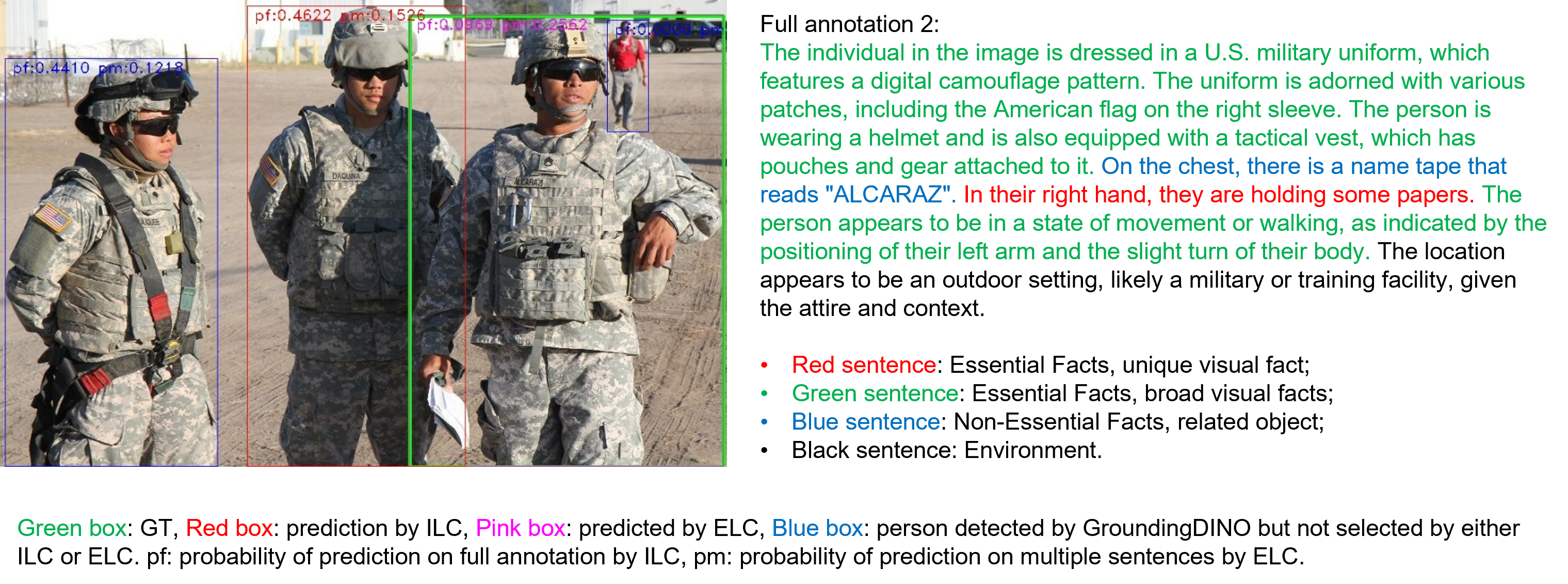}
\includegraphics[width=1.0\columnwidth]{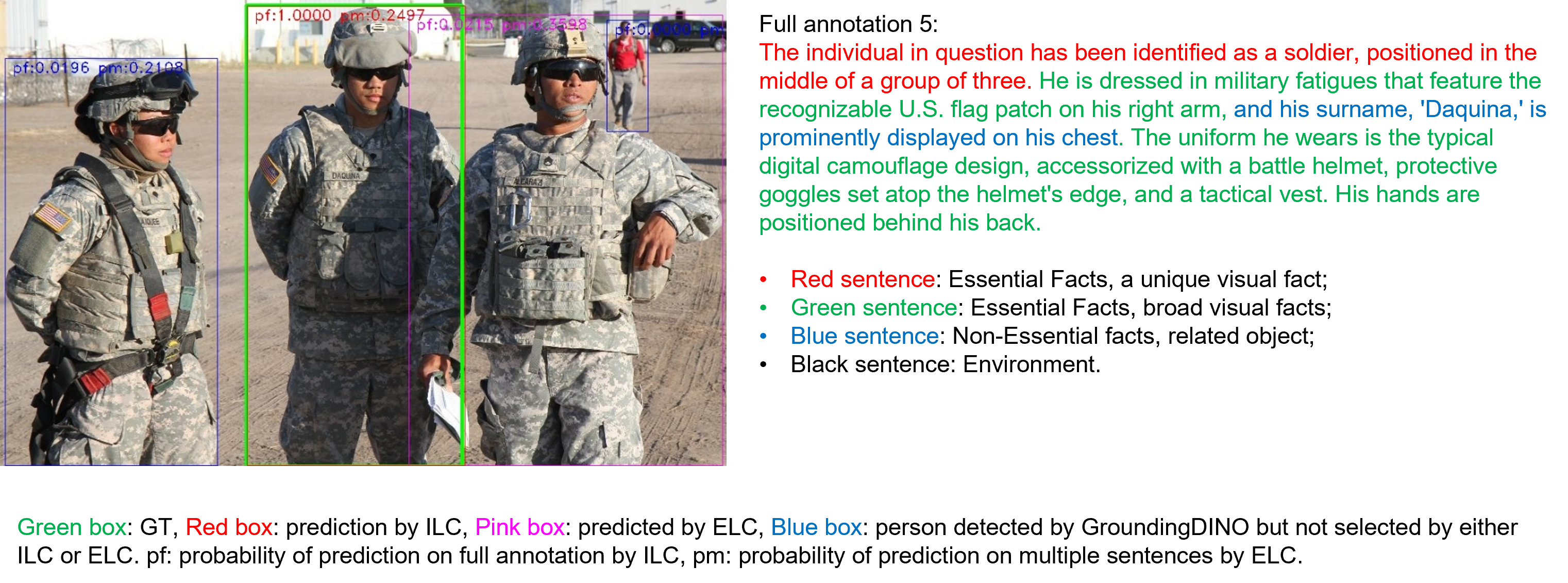}
\caption{Inconsistent examples from HC-RefLoCo.}
\label{Fig:Examples_HC-RefLoCo}
\end{figure*}

\subsection{Examples from NegBench}

Four inconsistent examples from NegBench are presented in Figure~\ref{Fig:Examples_NegBench}, where ILC on InternVL2-8B and ELC on factual and counterfactual reasoning produce inconsistent predictions. In each image, the red box and text indicates a pos object, and the blue box and text indicates a neg object. The text at the top of the image indicates that the referred object is not detected, with the red text referring to pos object and blue text for neg object. Under each image, the four text choices are presented. Below them, the blue texts present the GT choice, the choices predicted by ILC and ELC, respectively. At the bottom, the red text indicates the cause of the error.

In the first example, both GT choice and ILC prediction are not correct as they are based on the hallucination of non-existent object `carrot'. In ELC, the VFM (GroundingDINO) does not detect carrot on the table, leading to counterfactual inference and selecting the right choice. In the second image, both bus and car are partially occluded. Again, both GT choice and ILC prediction are not correct as they missed the presence of the partially occluded bus. However, in ELC, the VFM (GroundingDINO) detects the partially occluded bus in the left, which leads to logic inference on factual evidence to make the correct choice. In the third example, there is an ambiguous concept of `dining table'.
In the image, a man is using the small table at the back of the seat in front as dining table. On GT choice, the small table is not considered as dining table by human annotator. In ELC, GroundingDINO detects it as a dining table, leading to the 3rd choice. The explicit indication of the pos object, \ie, the red box and text shown in the image, is very helpful for logic justification of its choice. In the fourth example, GroundingDINO detects the desk as a dining table when it is asked to ground a dining table. Such error may cause wrong prediction of ELC. With the explicit indication of pos object as factual evidence, it supports logic validation of the correct answer.

\subsection{Examples from HC-RefCOCOg}

Four inconsistent examples from HC-RefCOCOg are presented in Figure~\ref{Fig:Examples_HC-RefCOCOg}, where ILC and ELC on factual and relational reasoning produce inconsistent predictions. In each image, the green box indicates the GT box, the red box and text indicate the prediction made by ILC with the probability score, the pink box and text indicate the prediction made by ELC, and the blue box and text indicate the detected referred object associated with ELC's prediction. Under each image, the phrase of referring expression is presented. At the bottom, the explanation of the boxes in the image are presented.

In the first example, ILC fails to distinguish the tank top with white shirt and makes a wrong prediction. On the other hand, GroundingDINO distinguishes the tank top correctly, leading to the correct prediction by logic reasoning on the grounded object and spatial relation with the person. In the second example, it might be difficult to discriminate the `black and gray shirt' with dark blue shirt as well as the relative length of hair. ILC makes a wrong prediction. However, GroundingDINO successfully grounds the `black and gray shirt' and `longer brown hair' in the image. Hence, ELC makes the correct prediction on the explicit visual evidence. The last two examples come from the same image. It is found that both ILC and ELC struggle to discriminate `blue color shirt' with `baby blue shirt'. The decision on ILC's prediction is unexplainable. However, ELC gives explicit visual evidence on its decision. 
These examples show that with explicit facts and relations, it is easy for us to make logic validation and justification on the final predictions on HC-REC task. 

\subsection{Examples from HC-RefLoCo}

Four inconsistent examples from HC-RefLoCo are presented in Figure~\ref{Fig:Examples_HC-RefLoCo}, where ILC with full annotation and ELC on InternVL2-8B with multiple sentences produce inconsistent predictions. In each image, the green box indicates the GT box, the red box and text indicates the prediction made by ILC with its probability, the pink box and text indicates the prediction made by ELC, and the blue box and text indicates other detected persons. On the right side of the image, the long context full annotation is presented. As indicated below the full annotation, on the categories of the sentences, the green sentence could be assumed as an Essential Fact with broad visual cues, the blue sentence could be assumed as a Non-Essential Fact, the red sentence could be considered as an Essential Fact with unique visual cue, and the black sentence can be assumed as an Environment description. At the bottom, the explanations of the boxes in the image are presented.

In the two examples, the green sentences would lead the attention to any of the three US soldiers. Hence, they could be considered as a sufficient cause to find the target person in the image, but not the necessary causes. The red sentences contain distinctive descriptions which would lead to the target person uniquely, and can be considered as a sufficient and necessary cause. The blue sentences contain the unique object(s) associated to the target person. Hence, it can be assumed as a necessary but not sufficient cause to localize the target person. The black sentences are usually less related to the target person in the HC-REC task. Hence, they can be considered as neither necessary nor sufficient conditions. 

In the first example, the ILC with full annotation as text input fails to localize the right person, as indicated by the red box. It may be confused by the green sentences and lose the cues for the correct prediction. However, the ELC on InternVL2-8B makes prediction based on the weighted sum of sentence categories, is able to correctly localize the target person, showing the effectiveness of logic reasoning on sentences for HC-REC task. In the second example, the first red sentence is not very effective in our experiment as the person bounding box is cropped and fed to VLM. The spatial cues may be lost. Hence, for the prediction based on split sentences, the sufficient and necessary conditions are weak. While on the full annotation, ILC makes the correct prediction, which maybe less affected by the truncation of the relatively shorter full annotation. The ELC on sentences fails to localize the correct person, due to the lack of sentences with both sufficient and necessary conditions, except for the first sentence with spatial cue information. However, it makes the second highest prediction on the correct person. On these two examples, the integrated predictions with aligned fusion produce the correct predictions, by fusing and complementing the strengths from both ILC and ELC.



%
%

\bibliographystyle{splncs04}
\bibliography{main}

@String(CVPR  = {IEEE Conf. Comput. Vis. Pattern Recog.})

@String(ICCV  = {Int. Conf. Comput. Vis.})

@String(ECCV  = {Eur. Conf. Comput. Vis.})

@String(AAAI  = {AAAI})

@String(CVPR  = {CVPR})

@String(ICCV  = {ICCV})

@String(ECCV  = {ECCV})

@inproceedings{alhamoud2025vision,
  author  = "Alhamoud, Kumail and Alshammari, Shaden and Tian, Yonglong and Li, Guohao and Torr, Philip and Kim, Yoon and Ghassemi, Marzyeh",
  title   = "{Vision-Language Models Do Not Understand Negation}",
  booktitle = "CVPR-25",
  year    = 2025,
}

@article{Balsdon:20,
  author  = "Balsdon, T. and Wyart, V. and Mamassian, P.",
  year    = 2020,
  title   = "{Confidence controls perceptual evidence accumulation}",
  journal = "Nat Commun",
  volume  = 11,
  number  = 1753,
doi = {https://doi.org/10.1038/s41467-020-15561-w},
}

@article{gemma_2025,
    title={Gemma 3},
    url={https://goo.gle/Gemma3Report},
    publisher={Kaggle},
    author={Gemma Team},
    year={2025}
}

@incollection{Bradley2012-BRADTA-6,
	author = "Richard Bradley",
	booktitle = "{Introduction to Formal Philosophy}",
	editor = "Sven Ove Hansson and Vincent F. Hendricks",
	pages = "611--655",
	title = "{Decision Theory: A Formal Philosophical Introduction}",
	year = "2018",
        publisher = "Springer International Publishing",
}

@inproceedings{NEURIPS2024_80f0cd03,
 author = "Wei, Fangyun and Zhao, Jinjing and Yan, Kun and Zhang, Hongyang and Xu, Chang",
 booktitle = "Advances in Neural Information Processing Systems",
 editor = "A. Globerson and L. Mackey and D. Belgrave and A. Fan and U. Paquet and J. Tomczak and C. Zhang",
 pages = "69566--69587",
 publisher = "Curran Associates, Inc.",
 title = "{A Large-Scale Human-Centric Benchmark for Referring Expression Comprehension in the LMM Era}",
 volume = "37",
 year = "2024",
}

@inproceedings{han2024zerorec,
  title={Zero-shot referring expression comprehension via structural similarity between images and captions},
  author={Han, Zeyu and Zhu, Fangrui and Lao, Qianru and Jiang, Huaizu},
  booktitle={Proceedings of the IEEE/CVF Conference on Computer Vision and Pattern Recognition},
  pages={14364--14374},
  year={2024}
}

@inproceedings{shen2024groundvlp,
  title={Groundvlp: Harnessing zero-shot visual grounding from vision-language pre-training and open-vocabulary object detection},
  author={Shen, Haozhan and Zhao, Tiancheng and Zhu, Mingwei and Yin, Jianwei},
  booktitle={Proceedings of the AAAI Conference on Artificial Intelligence},
  volume={38},
  number={5},
  pages={4766--4775},
  year={2024}
}

@misc{jiang2023mistral7b,
      title="{Mistral 7B}", 
      author="Albert Q. Jiang and Alexandre Sablayrolles and Arthur Mensch and Chris Bamford and Devendra Singh Chaplot and Diego de las Casas and Florian Bressand and Gianna Lengyel and Guillaume Lample and Lucile Saulnier and Lélio Renard Lavaud and Marie-Anne Lachaux and Pierre Stock and Teven Le Scao and Thibaut Lavril and Thomas Wang and Timothée Lacroix and William El Sayed",
      year="2023",
      eprint="2310.06825",
      archivePrefix="arXiv",
      primaryClass="cs.CL",    url="https://arxiv.org/abs/2310.06825", 
}

@InProceedings{Li_2025_CVPR,
    author    = "Li, Zongxia and Wu, Xiyang and Du, Hongyang and Liu, Fuxiao and Nghiem, Huy and Shi, Guangyao",
    title     = "{A Survey of State of the Art Large Vision Language Models: Benchmark Evaluations and Challenges}",
    booktitle = "Proceedings of the Computer Vision and Pattern Recognition Conference (CVPR) Workshops",
    month     = "June",
    year      = "2025",
    pages     = "1587-1606",
}

@misc{dang2024explainableinterpretablemultimodallarge,
      title="{Explainable and Interpretable Multimodal Large Language Models: A Comprehensive Survey}", 
      author="Yunkai Dang and Kaichen Huang and Jiahao Huo and Yibo Yan and Sirui Huang and Dongrui Liu and Mengxi Gao and Jie Zhang and Chen Qian and Kun Wang and Yong Liu and Jing Shao and Hui Xiong and Xuming Hu",
      year="2024",
      eprint="2412.02104",
      archivePrefix="arXiv",
      url="https://arxiv.org/abs/2412.02104", 
}

@inproceedings{NEURIPS2024_1e69ff56,
 author = "Li, Baiqi and Lin, Zhiqiu and Peng, Wenxuan and Nyandwi, Jean de Dieu and Jiang, Daniel and Ma, Zixian and Khanuja, Simran and Krishna, Ranjay and Neubig, Graham and Ramanan, Deva",
 booktitle = "{Advances in Neural Information Processing Systems}",
 pages = "17044--17068",
 publisher = "Curran Associates, Inc.",
 title = "NaturalBench: Evaluating Vision-Language Models on Natural Adversarial Samples",
 url = "https://proceedings.neurips.cc/paper_files/paper/2024/file/1e69ff56d0ebff0752ff29caaddc25dd-Paper-Datasets_and_Benchmarks_Track.pdf",
 volume = "37",
 year = "2024",
}

@misc{fu2024mmesurveycomprehensivesurveyevaluation,
      title="{MME-Survey: A Comprehensive Survey on Evaluation of Multimodal LLMs}", 
      author="Chaoyou Fu and Yi-Fan Zhang and Shukang Yin and Bo Li and Xinyu Fang and Sirui Zhao and Haodong Duan and Xing Sun and Ziwei Liu and Liang Wang and Caifeng Shan and Ran He",
      year="2024",
      eprint="2411.15296",
      archivePrefix="arXiv",
      primaryClass="cs.CV",
      url="https://arxiv.org/abs/2411.15296", 
}

@InProceedings{Li_2025_CVPRW,
    author    = "Li, Zongxia and Wu, Xiyang and Du, Hongyang and Liu, Fuxiao and Nghiem, Huy and Shi, Guangyao",
    title     = "{A Survey of State of the Art Large Vision Language Models: Alignment, Benchmark, Evaluations and Challenges}",
    booktitle = "Proceedings of the Computer Vision and Pattern Recognition Conference (CVPR) Workshops",
    month     = "June",
    year      = "2025",
    pages     = "1587-1606",
}

@misc{cheng2025empoweringllmslogicalreasoning,
      title="{Empowering LLMs with Logical Reasoning: A Comprehensive Survey}", 
      author="Fengxiang Cheng and Haoxuan Li and Fenrong Liu and Robert van Rooij and Kun Zhang and Zhouchen Lin",
      year="2025",
      eprint="2502.15652",
      archivePrefix="arXiv",
      primaryClass="cs.AI",
    url="https://arxiv.org/abs/2502.15652", 
}

@article{liu2023llava,
  title={Visual instruction tuning},
  author={Liu, Haotian and Li, Chunyuan and Wu, Qingyang and Lee, Yong Jae},
  journal={Advances in neural information processing systems},
  volume={36},
  pages={34892--34916},
  year={2023}
}

@misc{bai2023qwenvl,
      title={Qwen-VL: A Versatile Vision-Language Model for Understanding, Localization, Text Reading, and Beyond}, 
      author={Jinze Bai and Shuai Bai and Shusheng Yang and Shijie Wang and Sinan Tan and Peng Wang and Junyang Lin and Chang Zhou and Jingren Zhou},
      year={2023},
      eprint={2308.12966},
      archivePrefix={arXiv},
}

@article{awais2023amber,
  title={AMBER: advancing multimodal brain-computer interfaces for enhanced robustness—A dataset for naturalistic settings},
  author={Muhammad Ahsan Awais 
Muhammad Ahsan Awais, Peter Redmond, Tomas Emmanuel Ward and Graham Healy},
  journal={Front. Neuroergonomics},
  volume={4},
  doi={10.3389/fnrgo.2023.1216440},
  year={2023}
}

@inproceedings{feng2025bird,
    title="{{BIRD}: A Trustworthy Bayesian Inference Framework for Large Language Models}",
    author="Yu Feng and Ben Zhou and Weidong Lin and Dan Roth",
    booktitle="The Thirteenth International Conference on Learning Representations",
    year="2025",
    url="https://openreview.net/forum?id=fAAaT826Vv",
}

@InProceedings{Chen_2024_CVPR_InternVL,
    author    = "Chen, Zhe and Wu, Jiannan and Wang, Wenhai and Su, Weijie and Chen, Guo and Xing, Sen and Zhong, Muyan and Zhang, Qinglong and Zhu, Xizhou and Lu, Lewei and Li, Bin and Luo, Ping and Lu, Tong and Qiao, Yu and Dai, Jifeng",
    title     = "{InternVL: Scaling up Vision Foundation Models and Aligning for Generic Visual-Linguistic Tasks}",
    booktitle = "Proceedings of the IEEE/CVF Conference on Computer Vision and Pattern Recognition (CVPR)",
    month     = "June",
    year      = "2024",
    pages     = "24185-24198",
}

@misc{bai2023qwenvlversatilevisionlanguagemodel,
      title="{Qwen-VL: A Versatile Vision-Language Model for Understanding, Localization, Text Reading, and Beyond}", 
      author="Jinze Bai and Shuai Bai and Shusheng Yang and Shijie Wang and Sinan Tan and Peng Wang and Junyang Lin and Chang Zhou and Jingren Zhou",
      year="2023",
      eprint="2308.12966",
      archivePrefix="arXiv",
      primaryClass="cs.CV",
      url="https://arxiv.org/abs/2308.12966", 
}

@inproceedings{olausson-etal-2023-linc,
    title = "{{LINC}: A Neurosymbolic Approach for Logical Reasoning by Combining Language Models with First-Order Logic Provers}",
    author = "Olausson, Theo  and
      Gu, Alex  and
      Lipkin, Ben  and
      Zhang, Cedegao  and
      Solar-Lezama, Armando  and
      Tenenbaum, Joshua  and
      Levy, Roger",
    editor = "Bouamor, Houda  and
      Pino, Juan  and
      Bali, Kalika",
    booktitle = "Proceedings of the 2023 Conference on Empirical Methods in Natural Language Processing",
    month = dec,
    year = "2023",
    address = "Singapore",
    publisher = "Association for Computational Linguistics",
    url = "https://aclanthology.org/2023.emnlp-main.313/",
    doi = "10.18653/v1/2023.emnlp-main.313",
    pages = "5153--5176",
}

@inproceedings{feng-etal-2024-language,
    title = "{Language Models can be Deductive Solvers}",
    author = "Feng, Jiazhan  and
      Xu, Ruochen  and
      Hao, Junheng  and
      Sharma, Hiteshi  and
      Shen, Yelong  and
      Zhao, Dongyan  and
      Chen, Weizhu",
    editor = "Duh, Kevin  and
      Gomez, Helena  and
      Bethard, Steven",
    booktitle = "Findings of the Association for Computational Linguistics: NAACL 2024",
    month = "June",
    year = "2024",
    address = "Mexico City, Mexico",
    publisher = "Association for Computational Linguistics",
    url = "https://aclanthology.org/2024.findings-naacl.254/",
    doi = "10.18653/v1/2024.findings-naacl.254",
    pages = "4026--4042",
}

@inproceedings{calanzone2025logically,
    title="{Logically Consistent Language Models via Neuro-Symbolic Integration}",
    author="Diego, Calanzone and Stefano, Teso and Antonio, Vergari",
    booktitle="The Thirteenth International Conference on Learning Representations",
    year="2025",
    url="https://openreview.net/forum?id=7PGluppo4k",
}

@inproceedings{kirillov2023segany,
  title={Segment Anything},
  author={Kirillov, Alexander and Mintun, Eric and Ravi, Nikhila and Mao, Hanzi and Rolland, Chloe and Gustafson, Laura and Xiao, Tete and Whitehead, Spencer and Berg, Alexander C. and Lo, Wan-Yen and Doll{\'a}r, Piotr and Girshick, Ross},
  booktitle="International Conference on Learning Representations",
  year={2023},
}

@misc{zhao2023automaticmodelselectionlarge,
      title={Automatic Model Selection with Large Language Models for Reasoning}, 
      author={James Xu Zhao and Yuxi Xie and Kenji Kawaguchi and Junxian He and Michael Qizhe Xie},
      year={2023},
      eprint={2305.14333},
      archivePrefix={arXiv},
      primaryClass={cs.CL},
      url={https://arxiv.org/abs/2305.14333}, 
}

@misc{zhao2021calibrateuseimprovingfewshot,
      title={Calibrate Before Use: Improving Few-Shot Performance of Language Models}, 
      author={Tony Z. Zhao and Eric Wallace and Shi Feng and Dan Klein and Sameer Singh},
      year={2021},
      eprint={2102.09690},
      archivePrefix={arXiv},
      primaryClass={cs.CL},
      url={https://arxiv.org/abs/2102.09690}, 
}

@misc{wang2023selfconsistencyimproveschainthought,
      title={Self-Consistency Improves Chain of Thought Reasoning in Language Models}, 
      author={Xuezhi Wang and Jason Wei and Dale Schuurmans and Quoc Le and Ed Chi and Sharan Narang and Aakanksha Chowdhery and Denny Zhou},
      year={2023},
      eprint={2203.11171},
      archivePrefix={arXiv},
      primaryClass={cs.CL},
      url={https://arxiv.org/abs/2203.11171}, 
}

@misc{zhao2025surveylargelanguagemodels,
      title="{A Survey of Large Language Models}", 
      author="Wayne Xin Zhao and Kun Zhou and Junyi Li and Tianyi Tang and Xiaolei Wang and Yupeng Hou and Yingqian Min and Beichen Zhang and Junjie Zhang and Zican Dong and Yifan Du and Chen Yang and Yushuo Chen and Zhipeng Chen and Jinhao Jiang and Ruiyang Ren and Yifan Li and Xinyu Tang and Zikang Liu and Peiyu Liu and Jian-Yun Nie and Ji-Rong Wen",
      year="2025",
      eprint="2303.18223",
      archivePrefix="arXiv",
      primaryClass="cs.CL",
      url="https://arxiv.org/abs/2303.18223", 
}

@misc{kil2025mllmcompbenchcomparativereasoningbenchmark,
      title={MLLM-CompBench: A Comparative Reasoning Benchmark for Multimodal LLMs}, 
      author={Jihyung Kil and Zheda Mai and Justin Lee and Zihe Wang and Kerrie Cheng and Lemeng Wang and Ye Liu and Arpita Chowdhury and Wei-Lun Chao},
      year={2025},
      eprint={2407.16837},
      archivePrefix={arXiv},
      primaryClass={cs.CV},
      url={https://arxiv.org/abs/2407.16837}, 
}

@misc{fu2025mmecomprehensiveevaluationbenchmark,
      title={MME: A Comprehensive Evaluation Benchmark for Multimodal Large Language Models}, 
      author={Chaoyou Fu and Peixian Chen and Yunhang Shen and Yulei Qin and Mengdan Zhang and Xu Lin and Jinrui Yang and Xiawu Zheng and Ke Li and Xing Sun and Yunsheng Wu and Rongrong Ji and Caifeng Shan and Ran He},
      year={2025},
      eprint={2306.13394},
      archivePrefix={arXiv},
      primaryClass={cs.CV},
      url={https://arxiv.org/abs/2306.13394}, 
}

@inproceedings{yue2024mmmu,
  title={Mmmu: A massive multi-discipline multimodal understanding and reasoning benchmark for expert agi},
  author={Yue, Xiang and Ni, Yuansheng and Zhang, Kai and Zheng, Tianyu and Liu, Ruoqi and Zhang, Ge and Stevens, Samuel and Jiang, Dongfu and Ren, Weiming and Sun, Yuxuan and others},
  booktitle={Proceedings of the IEEE/CVF Conference on Computer Vision and Pattern Recognition},
  pages={9556--9567},
  year={2024}
}

@misc{li2024surveybenchmarksmultimodallarge,
      title={A Survey on Benchmarks of Multimodal Large Language Models}, 
      author={Jian Li and Weiheng Lu and Hao Fei and Meng Luo and Ming Dai and Min Xia and Yizhang Jin and Zhenye Gan and Ding Qi and Chaoyou Fu and Ying Tai and Wankou Yang and Yabiao Wang and Chengjie Wang},
      year={2024},
      eprint={2408.08632},
      archivePrefix={arXiv},
      primaryClass={cs.CL},
      url={https://arxiv.org/abs/2408.08632}, 
}

@misc{wang2024comprehensivereviewmultimodallarge,
      title={A Comprehensive Review of Multimodal Large Language Models: Performance and Challenges Across Different Tasks}, 
      author={Jiaqi Wang and Hanqi Jiang and Yiheng Liu and Chong Ma and Xu Zhang and Yi Pan and Mengyuan Liu and Peiran Gu and Sichen Xia and Wenjun Li and Yutong Zhang and Zihao Wu and Zhengliang Liu and Tianyang Zhong and Bao Ge and Tuo Zhang and Ning Qiang and Xintao Hu and Xi Jiang and Xin Zhang and Wei Zhang and Dinggang Shen and Tianming Liu and Shu Zhang},
      year={2024},
      eprint={2408.01319},
      archivePrefix={arXiv},
      primaryClass={cs.AI},
      url={https://arxiv.org/abs/2408.01319}, 
}

@misc{liu2024mmbenchmultimodalmodelallaround,
      title={MMBench: Is Your Multi-modal Model an All-around Player?}, 
      author={Yuan Liu and Haodong Duan and Yuanhan Zhang and Bo Li and Songyang Zhang and Wangbo Zhao and Yike Yuan and Jiaqi Wang and Conghui He and Ziwei Liu and Kai Chen and Dahua Lin},
      year={2024},
      eprint={2307.06281},
      archivePrefix={arXiv},
      primaryClass={cs.CV},
      url={https://arxiv.org/abs/2307.06281}, 
}

@misc{geminiteam2025geminifamilyhighlycapable,
      title="{Gemini: A Family of Highly Capable Multimodal Models}", 
      author="{Gemini Team}",
      year="2025",
      eprint="2312.11805",
      archivePrefix="arXiv",
      primaryClass="cs.CL",
      url="https://arxiv.org/abs/2312.11805", 
}

@misc{yang2023dawnlmmspreliminaryexplorations,
      title="{The Dawn of LMMs: Preliminary Explorations with GPT-4V(ision)}", 
      author="Zhengyuan Yang and Linjie Li and Kevin Lin and Jianfeng Wang and Chung-Ching Lin and Zicheng Liu and Lijuan Wang",
      year="2023",
      eprint="2309.17421",
      archivePrefix="arXiv",
      primaryClass="cs.CV",
      url="https://arxiv.org/abs/2309.17421", 
}

@misc{liu2023visualinstructiontuning,
      title="{Visual Instruction Tuning}", 
      author="Haotian Liu and Chunyuan Li and Qingyang Wu and Yong Jae Lee",
      year="2023",
      eprint="2304.08485",
      archivePrefix="arXiv",
      primaryClass="cs.CV",
      url="https://arxiv.org/abs/2304.08485", 
}

@article{Huang_2025,
   title={A Survey on Hallucination in Large Language Models: Principles, Taxonomy, Challenges, and Open Questions},
   volume={43},
   ISSN={1558-2868},
   url={http://dx.doi.org/10.1145/3703155},
   DOI={10.1145/3703155},
   number={2},
   journal={ACM Transactions on Information Systems},
   publisher={Association for Computing Machinery (ACM)},
   author={Huang, Lei and Yu, Weijiang and Ma, Weitao and Zhong, Weihong and Feng, Zhangyin and Wang, Haotian and Chen, Qianglong and Peng, Weihua and Feng, Xiaocheng and Qin, Bing and Liu, Ting},
   year={2025},
   month=jan, pages={1–55} }

@misc{yuan2025mmereasoningcomprehensivebenchmarklogical,
      title={MME-Reasoning: A Comprehensive Benchmark for Logical Reasoning in MLLMs}, 
      author={Jiakang Yuan and Tianshuo Peng and Yilei Jiang and Yiting Lu and Renrui Zhang and Kaituo Feng and Chaoyou Fu and Tao Chen and Lei Bai and Bo Zhang and Xiangyu Yue},
      year={2025},
      eprint={2505.21327},
      archivePrefix={arXiv},
      primaryClass={cs.AI},
      url={https://arxiv.org/abs/2505.21327}, 
}

@article{Rahman_2026,
   title={Hallucination to truth: a review of fact-checking and factuality evaluation in large language models},
   volume={59},
   ISSN={1573-7462},
   url={http://dx.doi.org/10.1007/s10462-025-11454-w},
   DOI={10.1007/s10462-025-11454-w},
   number={2},
   journal={Artificial Intelligence Review},
   publisher={Springer Science and Business Media LLC},
   author={Rahman, Subhey Sadi and Islam, Md. Adnanul and Alam, Md. Mahbub and Zeba, Musarrat and Rahman, Md. Abdur and Chowa, Sadia Sultana and Raiaan, Mohaimenul Azam Khan and Azam, Sami},
   year={2026},
   month=jan }

@misc{wang2024factualitylargelanguagemodels,
      title={Factuality of Large Language Models: A Survey}, 
      author={Yuxia Wang and Minghan Wang and Muhammad Arslan Manzoor and Fei Liu and Georgi Georgiev and Rocktim Jyoti Das and Preslav Nakov},
      year={2024},
      eprint={2402.02420},
      archivePrefix={arXiv},
      primaryClass={cs.CL},
      url={https://arxiv.org/abs/2402.02420}, 
}

@misc{bai2025hallucinationmultimodallargelanguage,
      title={Hallucination of Multimodal Large Language Models: A Survey}, 
      author={Zechen Bai and Pichao Wang and Tianjun Xiao and Tong He and Zongbo Han and Zheng Zhang and Mike Zheng Shou},
      year={2025},
      eprint={2404.18930},
      archivePrefix={arXiv},
      primaryClass={cs.CV},
      url={https://arxiv.org/abs/2404.18930}, 
}

@InProceedings{pmlr-v139-radford21a,
  title = 	 "{Learning Transferable Visual Models From Natural Language Supervision}",
  author =       "Radford, Alec and Kim, Jong Wook and Hallacy, Chris and Ramesh, Aditya and Goh, Gabriel and Agarwal, Sandhini and Sastry, Girish and Askell, Amanda and Mishkin, Pamela and Clark, Jack and Krueger, Gretchen and Sutskever, Ilya",
  booktitle = 	 "Proceedings of the 38th International Conference on Machine Learning",
  pages = 	 "8748--8763",
  year = 	 "2021",
  editor = 	 "Meila, Marina and Zhang, Tong",
  volume = 	 "139",
  series = 	 "Proceedings of Machine Learning Research",
  publisher =    "PMLR",
  url = 	 "https://proceedings.mlr.press/v139/radford21a.html",
}

@InProceedings{pmlr-v162-li22n,
  title = 	 "{{BLIP}: Bootstrapping Language-Image Pre-training for Unified Vision-Language Understanding and Generation}",
  author =       "Li, Junnan and Li, Dongxu and Xiong, Caiming and Hoi, Steven",
  booktitle = 	 "Proceedings of the 39th International Conference on Machine Learning",
  pages = 	 "12888--12900",
  year = 	 "2022",
  editor = 	 "Chaudhuri, Kamalika and Jegelka, Stefanie and Song, Le and Szepesvari, Csaba and Niu, Gang and Sabato, Sivan",
  volume = 	 "162",
  series = 	 "Proceedings of Machine Learning Research",
  publisher =    "PMLR",
  url = 	 "https://proceedings.mlr.press/v162/li22n.html",
}

@misc{jia2021scalingvisualvisionlanguagerepresentation,
      title="{Scaling Up Visual and Vision-Language Representation Learning With Noisy Text Supervision}", 
      author="Chao Jia and Yinfei Yang and Ye Xia and Yi-Ting Chen and Zarana Parekh and Hieu Pham and Quoc V. Le and Yunhsuan Sung and Zhen Li and Tom Duerig",
      year="2021",
      eprint="2102.05918",
      archivePrefix="arXiv",
      primaryClass="cs.CV",
      url="https://arxiv.org/abs/2102.05918", 
}

@article{Yin_2024,
   title="{A survey on multimodal large language models}",
   volume="11",
   ISSN="2053-714X",
   url="http://dx.doi.org/10.1093/nsr/nwae403",
   DOI="10.1093/nsr/nwae403",
   number="12",
   journal="National Science Review",
   publisher="Oxford University Press (OUP)",
   author="Yin, Shukang and Fu, Chaoyou and Zhao, Sirui and Li, Ke and Sun, Xing and Xu, Tong and Chen, Enhong",
   year="2024",
}

@ARTICLE{10.3389/frai.2022.806403,
AUTHOR="Wang, Pei  and Hahm, Christian  and Hammer, Patrick",       
TITLE="{A Model of Unified Perception and Cognition}",   
JOURNAL="Frontiers in Artificial Intelligence",  
VOLUME="Volume 5 - 2022",
YEAR="2022",
URL="https://www.frontiersin.org/journals/artificial-intelligence/articles/10.3389/frai.2022.806403",
DOI="10.3389/frai.2022.806403",
ISSN="2624-8212",
}

@ARTICLE{Yang2024_10234506,
  author="Yang, Xiaofeng and Liu, Fayao and Lin, Guosheng",
  journal="IEEE Transactions on Multimedia", 
  title="{Neural Logic Vision Language Explainer}", 
  year="2024",
  volume="26",
  pages="3331-3340",
}

@inproceedings{yu2016refcoco,
  title={Modeling context in referring expressions},
  author={Yu, Licheng and Poirson, Patrick and Yang, Shan and Berg, Alexander C and Berg, Tamara L},
  booktitle={European conference on computer vision},
  pages={69--85},
  year={2016},
  organization={Springer}
}

@inproceedings{mao2016refcocog,
  title={Generation and comprehension of unambiguous object descriptions},
  author={Mao, Junhua and Huang, Jonathan and Toshev, Alexander and Camburu, Oana and Yuille, Alan L and Murphy, Kevin},
  booktitle={Proceedings of the IEEE conference on computer vision and pattern recognition},
  pages={11--20},
  year={2016}
}

@misc{xiao2024visualgroundingsurvey,
      title="{Towards Visual Grounding: A Survey}", 
      author="Linhui Xiao and Xiaoshan Yang and Xiangyuan Lan and Yaowei Wang and Changsheng Xu",
      year="2024",
      eprint="2412.20206",
      archivePrefix="arXiv",
      primaryClass="cs.CV",
      url="https://arxiv.org/abs/2412.20206", 
}

@inproceedings{Khan_2024_CVPR,
    author    = {Khan, Zaid and Fu, Yun},
    title     = {Consistency and Uncertainty: Identifying Unreliable Responses From Black-Box Vision-Language Models for Selective Visual Question Answering},
    booktitle = {Proceedings of the IEEE/CVF Conference on Computer Vision and Pattern Recognition (CVPR)},
    month     = {June},
    year      = {2024},
    pages     = {10854-10863}
}

@ARTICLE{ZiyangLi2023,
  author="Li, Ziyang and Huang, Jiani and Naik, Mayur",
  journal="Proc. ACM Program. Lang", 
  title="{Scallop: A Language for Neurosymbolic Programming}", 
  year="2023",
  volume="7",
  pages="166:1-166:25",
}

@inproceedings{Huang2023LASERAN,
  title={LASER: A Neuro-Symbolic Framework for Learning Spatio-Temporal Scene Graphs with Weak Supervision},
  author={Jiani Huang and Ziyang Li and David Jacobs and M. Naik and Ser Nam Lim},
  booktitle={International Conference on Learning Representations},
  year={2025},
  url={https://api.semanticscholar.org/CorpusID:258179880}
}

@misc{pournemat2025reasoninguncertaintyexploringprobabilistic,
      title={Reasoning Under Uncertainty: Exploring Probabilistic Reasoning Capabilities of LLMs}, 
      author={Mobina Pournemat and Keivan Rezaei and Gaurang Sriramanan and Arman Zarei and Jiaxiang Fu and Yang Wang and Hamid Eghbalzadeh and Soheil Feizi},
      year={2025},
      eprint={2509.10739},
      archivePrefix={arXiv},
      primaryClass={cs.CL},
      url={https://arxiv.org/abs/2509.10739}, 
}

@article{Cheng2025TheFL,
  title={The FACTS Leaderboard: A Comprehensive Benchmark for Large Language Model Factuality},
  author={Aileen Cheng and Alon Jacovi and Amir Globerson and Ben Golan and Charles Kwong and Chris Alberti and Connie Tao and Eyal Ben-David and Gaurav Singh Tomar and Lukas Haas and Yonatan Bitton and Adam E. Bloniarz and Aijun Bai and Andrew Wang and Anfal Siddiqui and Arturo Bajuelos Castillo and Aviel Atias and Chang Liu and Corey Fry and Daniel Balle and Deepanway Ghosal and Doron Kukliansky and Dror Marcus and Elena Gribovskaya and Eran. O. Ofek and Honglei Zhuang and Itay Laish and Jan Ackermann and Lily Wang and Meg Risdal and Megan Barnes and Michael Fink and Mohamed Amin and Moran Ambar and Natan Potikha and Nikita Gupta and Nitzan Katz and Noam Velan and Ofir Roval and Ori Ram and Polina Zablotskaia and Prathamesh Bang and Priyanka Agrawal and Rakesh Ghiya and Sanjay Ganapathy and Simon Baumgartner and Sofia Erell and Sushant Prakash and Thibault Sellam and Vikram Rao and Xuanhui Wang and Yaroslav Akulov and Yulong Yang and Zhen-Bo Ou Yang and Zhixin Lai and Zhongru Wu and Anca Dragan and Avinatan Hassidim and Fernando Pereira and Slav Petrov and Srinivasan Venkatachary and Tulsee Doshi and Yossi Matias and Sasha Goldshtein and Dipanjan Das},
  journal={ArXiv},
  year={2025},
  volume={abs/2512.10791},
  url={https://api.semanticscholar.org/CorpusID:283737312}
}

@article{AKOGLU201891,
title = {User's guide to correlation coefficients},
journal = {Turkish Journal of Emergency Medicine},
volume = {18},
number = {3},
pages = {91-93},
year = {2018},
issn = {2452-2473},
doi = {https://doi.org/10.1016/j.tjem.2018.08.001},
url = {https://www.sciencedirect.com/science/article/pii/S2452247318302164},
author = {Haldun Akoglu}
}

@misc{qwen3technicalreport,
      title={Qwen3 Technical Report}, 
      author={Qwen Team},
      year={2025},
      eprint={2505.09388},
      archivePrefix={arXiv},
      primaryClass={cs.CL},
      url={https://arxiv.org/abs/2505.09388}, 
}

@article{EVA-CLIP-18B,
  title={EVA-CLIP-18B: Scaling CLIP to 18 Billion Parameters}, 
  author={Quan Sun and Jinsheng Wang and Qiying Yu and Yufeng Cui and Fan Zhang and Xiaosong Zhang and Xinlong Wang},
  journal={arXiv preprint arXiv:2402.04252},
  year={2023}
}

@INPROCEEDINGS{8953451,
  author={Hudson, Drew A. and Manning, Christopher D.},
  booktitle={2019 IEEE/CVF Conference on Computer Vision and Pattern Recognition (CVPR)}, 
  title={GQA: A New Dataset for Real-World Visual Reasoning and Compositional Question Answering}, 
  year={2019},
  volume={},
  number={},
  pages={6693-6702},
  doi={10.1109/CVPR.2019.00686}}

@InProceedings{Chen_2022_CVPR,
    author    = {Chen, Chongyan and Anjum, Samreen and Gurari, Danna},
    title     = {Grounding Answers for Visual Questions Asked by Visually Impaired People},
    booktitle = {Proceedings of the IEEE/CVF Conference on Computer Vision and Pattern Recognition (CVPR)},
    month     = {June},
    year      = {2022},
    pages     = {19098-19107}
}

@misc{reich2022visuallygroundedvqalatticebased,
      title={Visually Grounded VQA by Lattice-based Retrieval}, 
      author={Daniel Reich and Felix Putze and Tanja Schultz},
      year={2022},
      eprint={2211.08086},
      archivePrefix={arXiv},
      primaryClass={cs.CV},
      url={https://arxiv.org/abs/2211.08086}, 
}

@inproceedings{reich-etal-2023-measuring,
    title = "Measuring Faithful and Plausible Visual Grounding in {VQA}",
    author = "Reich, Daniel  and
      Putze, Felix  and
      Schultz, Tanja",
    editor = "Bouamor, Houda  and
      Pino, Juan  and
      Bali, Kalika",
    booktitle = "Findings of the Association for Computational Linguistics: EMNLP 2023",
    month = dec,
    year = "2023",
    address = "Singapore",
    publisher = "Association for Computational Linguistics",
    url = "https://aclanthology.org/2023.findings-emnlp.206/",
    doi = "10.18653/v1/2023.findings-emnlp.206",
    pages = "3129--3144",
}

@InProceedings{Xiao_2024_CVPR,
    author    = {Xiao, Junbin and Yao, Angela and Li, Yicong and Chua, Tat-Seng},
    title     = {Can I Trust Your Answer? Visually Grounded Video Question Answering},
    booktitle = {Proceedings of the IEEE/CVF Conference on Computer Vision and Pattern Recognition (CVPR)},
    month     = {June},
    year      = {2024},
    pages     = {13204-13214}
}

@article{DBLP:journals/corr/abs-2509-11862,
  author={Haodi Ma and Vyom Pathak and Daisy Zhe Wang},
  title={Bridging Vision Language Models and Symbolic Grounding for Video Question Answering},
  journal={CoRR},
  volume={abs/2509.11862},
  year={2025},
  url={https://doi.org/10.48550/arXiv.2509.11862},
  doi={10.48550/ARXIV.2509.11862},
  eprinttype={arXiv},
  eprint={2509.11862},
  bibsource={dblp computer science bibliography, https://dblp.org}
}

@inproceedings{di2024grounded_ego,
  title={Grounded question-answering in long egocentric videos},
  author={Di, Shangzhe and Xie, Weidi},
  booktitle={Proceedings of the IEEE/CVF Conference on Computer Vision and Pattern Recognition},
  pages={12934--12943},
  year={2024}
}

@INPROCEEDINGS{7780909,
  author={Zhu, Yuke and Groth, Oliver and Bernstein, Michael and Fei-Fei, Li},
  booktitle={2016 IEEE Conference on Computer Vision and Pattern Recognition (CVPR)}, 
  title={Visual7W: Grounded Question Answering in Images}, 
  year={2016},
  volume={},
  number={},
  pages={4995-5004},
  doi={10.1109/CVPR.2016.540}}

@inproceedings{li2022invariant,
  title={Invariant grounding for video question answering},
  author={Li, Yicong and Wang, Xiang and Xiao, Junbin and Ji, Wei and Chua, Tat-Seng},
  booktitle={Proceedings of the IEEE/CVF Conference on Computer Vision and Pattern Recognition},
  pages={2928--2937},
  year={2022}
}

@INPROCEEDINGS{9578483,
  author={Khan, Aisha Urooj and Kuehne, Hilde and Duarte, Kevin and Gan, Chuang and Lobo, Niels and Shah, Mubarak},
  booktitle={2021 IEEE/CVF Conference on Computer Vision and Pattern Recognition (CVPR)}, 
  title={Found a Reason for me? Weakly-supervised Grounded Visual Question Answering using Capsules}, 
  year={2021},
  volume={},
  number={},
  pages={8461-8470},
  doi={10.1109/CVPR46437.2021.00836}}

@InProceedings{10.1007/978-3-031-19833-5_38,
author="Khan, Aisha Urooj
and Kuehne, Hilde
and Gan, Chuang
and Lobo, Niels Da Vitoria
and Shah, Mubarak",
editor="Avidan, Shai
and Brostow, Gabriel
and Ciss{\'e}, Moustapha
and Farinella, Giovanni Maria
and Hassner, Tal",
title="Weakly Supervised Grounding for VQA in Vision-Language Transformers",
booktitle="Computer Vision -- ECCV 2022",
year="2022",
publisher="Springer Nature Switzerland",
address="Cham",
pages="652--670",
isbn="978-3-031-19833-5"
}

@article{chen2024lcv2,
  title={LCV2: a universal pretraining-free framework for grounded visual question answering},
  author={Chen, Yuhan and Su, Lumei and Chen, Lihua and Lin, Zhiwei},
  journal={Electronics},
  volume={13},
  number={11},
  pages={2061},
  year={2024},
  publisher={MDPI}
}

@inproceedings{chen2023vqatheraphy,
  title={Vqa therapy: Exploring answer differences by visually grounding answers},
  author={Chen, Chongyan and Anjum, Samreen and Gurari, Danna},
  booktitle={Proceedings of the IEEE/CVF International Conference on Computer Vision},
  pages={15315--15325},
  year={2023}
}

@article{malinin2020uncertainty,
  title={Uncertainty estimation in autoregressive structured prediction},
  author={Malinin, Andrey and Gales, Mark},
  journal={arXiv preprint arXiv:2002.07650},
  year={2020}
}

@misc{li2023scalloplanguageneurosymbolicprogramming,
      title={Scallop: A Language for Neurosymbolic Programming}, 
      author={Ziyang Li and Jiani Huang and Mayur Naik},
      year={2023},
      eprint={2304.04812},
      archivePrefix={arXiv},
      primaryClass={cs.PL},
      url={https://arxiv.org/abs/2304.04812}, 
}

@InProceedings{10.1007/978-3-031-72630-9_3,
author="Kamath, Amita
and Hsieh, Cheng-Yu
and Chang, Kai-Wei
and Krishna, Ranjay",
editor="Leonardis, Ale{\v{s}}
and Ricci, Elisa
and Roth, Stefan
and Russakovsky, Olga
and Sattler, Torsten
and Varol, G{\"u}l",
title="The Hard Positive Truth About Vision-Language Compositionality",
booktitle="Computer Vision -- ECCV 2024",
year="2025",
publisher="Springer Nature Switzerland",
address="Cham",
pages="37--54",
}

@article{Wang2023EquivariantSF,
  title={Equivariant Similarity for Vision-Language Foundation Models},
  author={Tan Wang and Kevin Lin and Linjie Li and Chung-Ching Lin and Zhengyuan Yang and Hanwang Zhang and Zicheng Liu and Lijuan Wang},
  journal={2023 IEEE/CVF International Conference on Computer Vision (ICCV)},
  year={2023},
  pages={11964-11974},
  url={https://api.semanticscholar.org/CorpusID:257766245}
}

@inproceedings{10.5555/3666122.3666201,
author = {Yarom, Michal and Bitton, Yonatan and Changpinyo, Soravit and Aharoni, Roee and Herzig, Jonathan and Lang, Oran and Ofek, Eran and Szpektor, Idan},
title = {What you see is what you read? improving text-image alignment evaluation},
year = {2023},
publisher = {Curran Associates Inc.},
address = {Red Hook, NY, USA},
booktitle = {Proceedings of the 37th International Conference on Neural Information Processing Systems},
articleno = {79},
numpages = {19},
location = {New Orleans, LA, USA},
series = {NIPS '23}
}

@article{zhi2025tfar,
title={{TFAR}: A Training-Free Framework for Autonomous Reliable Reasoning in Visual Question Answering},
author={Zhuo Zhi and Chen Feng and Adam Daneshmend and Mine Orlu and Andreas Demosthenous and Lu Yin and Da Li and Ziquan Liu and Miguel R. D. Rodrigues},
journal={Transactions on Machine Learning Research},
issn={2835-8856},
year={2025},
url={https://openreview.net/forum?id=cBAKeZN3jy},
note={}
}

@book{10.1093/oso/9780198537465.001.0001,
    author = {},
    editor = {Gabbay, Dov M and Hogger, C J and Robinson, J A and Siekmann, J},
    title = {Handbook of Logic in Artificial Intelligence and Logic Programming},
    publisher = {Oxford University Press},
    year = {1994},
    month = {03},
    isbn = {9780198537465},
    doi = {10.1093/oso/9780198537465.001.0001},
    url = {https://doi.org/10.1093/oso/9780198537465.001.0001},
}

@book{Huth_Ryan_2004, 
place={Cambridge}, 
edition={2}, 
title={Logic in Computer Science: Modelling and Reasoning about Systems}, 
publisher={Cambridge University Press}, 
author={Huth, Michael and Ryan, Mark}, 
year={2004}
}

\end{document}